\pdfoutput=1
\documentclass{article}

\PassOptionsToPackage{numbers,compress}{natbib}
\usepackage[main, final]{neurips_2026}
\renewcommand{\acksection}{\section*{Acknowledgments}}
\usepackage[utf8]{inputenc}
\usepackage[T1]{fontenc}
\usepackage{amsmath,amssymb,amsfonts,booktabs,graphicx,microtype,multirow,placeins,float}
\usepackage{xcolor}
\usepackage[ruled,vlined,linesnumbered]{algorithm2e}
\usepackage{tikz}
\usetikzlibrary{arrows.meta,positioning,calc,fit,backgrounds}
\usepackage{hyperref}
\usepackage{url}

\title{SparDA: Sparse Decoupled Attention for Efficient Long-Context LLM Inference}
\author{%
  Yaosheng Fu\textsuperscript{1},
  Guangxuan Xiao\textsuperscript{2}\thanks{Work done while at NVIDIA.},
  Xin Dong\textsuperscript{3}\footnotemark[1],
  Song Han\textsuperscript{1,4},
  Oreste Villa\textsuperscript{1} \\
  \textsuperscript{1}NVIDIA\quad
  \textsuperscript{2}Thinking Machines Lab\quad
  \textsuperscript{3}ByteDance Seed\quad
  \textsuperscript{4}MIT
}

\newcommand{\method}{SparDA}

\begin{document}
\maketitle

\begin{abstract}
Sparse attention reduces compute and memory bandwidth for long-context LLM inference. However, two key challenges remain: (1) KV cache capacity still grows with sequence length, and offloading to CPU memory introduces a PCIe transfer bottleneck; (2) sparse selection during prefill retains $O(T^2)$ complexity and can dominate attention cost at long contexts. We propose SparDA, a decoupled sparse attention architecture that introduces a fourth per-layer projection, the \emph{Forecast}, alongside Query, Key, and Value. The Forecast predicts the KV blocks needed by the next layer, enabling lookahead selection that overlaps CPU-to-GPU prefetch with current-layer execution. Because Forecast is decoupled from the attention query, our GQA implementation uses one Forecast head per GQA group, reducing selection overhead versus the original multi-head selector. On the evaluated 8B models, SparDA adds $<$0.5\% parameters and trains only the Forecast projections using attention-derived supervision from the frozen backbone. On two sparse-pretrained 8B models, SparDA maintains accuracy comparable to the full-query sparse-attention baseline while delivering up to 1.25$\times$ prefill speedup and 1.7$\times$ decode speedup over the sparse-attention offload baseline. By enabling larger feasible batch sizes on a single GPU, SparDA further reaches up to 5.3$\times$ higher decode throughput than the non-offload sparse baseline on MiniCPM4.1-8B. Our adaptation to GLM-5.1 (754B MoE) also achieves accuracy comparable to native token-level DSA, extending the accuracy evaluation beyond block-sparse 8B models. Our source code is available at \url{https://github.com/NVlabs/SparDA}.
\end{abstract}

\begin{figure}[t]
\centering
\includegraphics[width=0.95\textwidth]{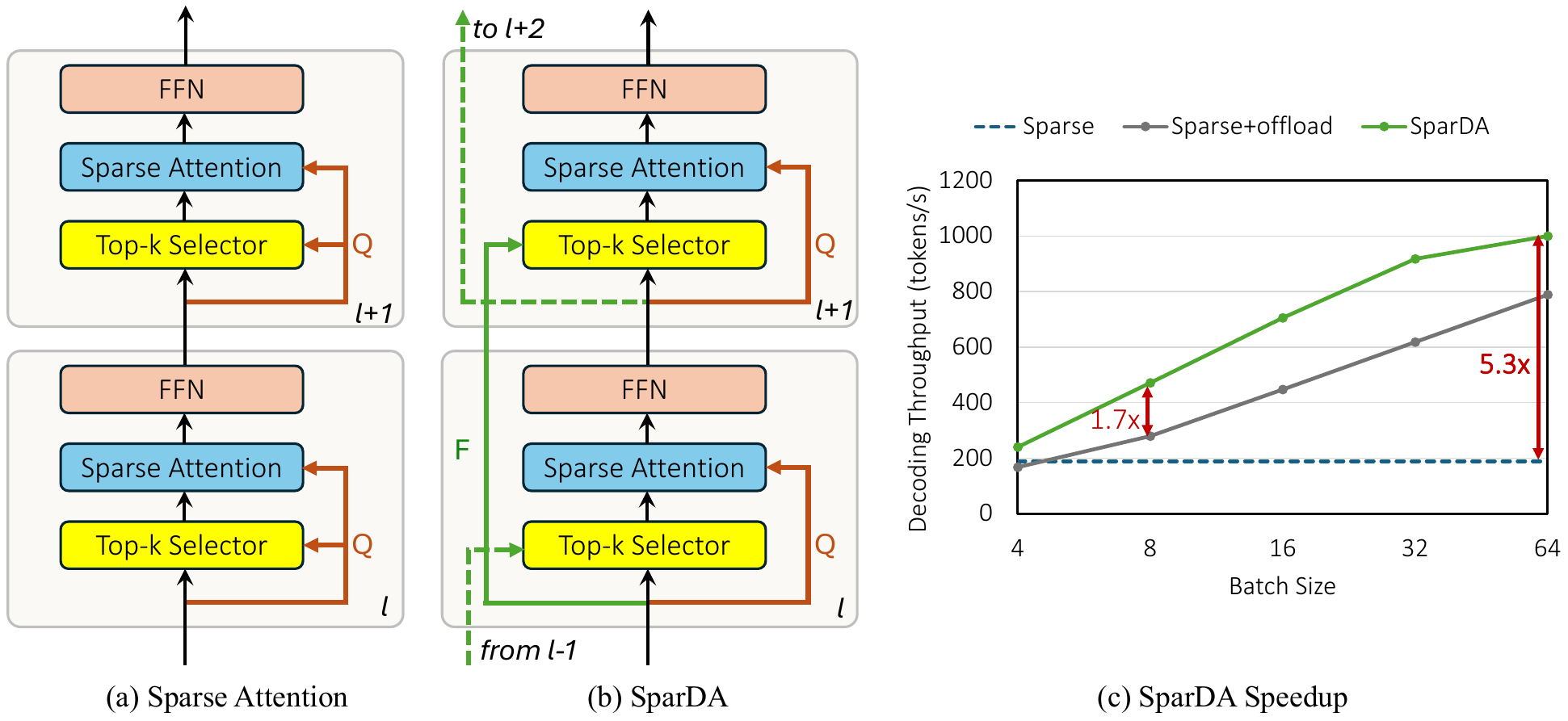}
\caption{Overview of \method. The Forecast $\mathbf{F}_l$ drives top-$k$ selection for layer $l{+}1$, while $\mathbf{Q}_l$ performs sparse attention, decoupling selection from attention and exposing lookahead KV prefetch from CPU. On MiniCPM4.1-8B at 128K on H100, \method\ achieves up to 1.7$\times$ decode speedup over sparse offload and 5.3$\times$ higher throughput than non-offload sparse via larger feasible batches. The dashed line marks the non-offload sparse throughput at its largest feasible batch (B4).}
\label{fig:concept}
\end{figure}

\section{Introduction}
\label{sec:introduction}

Modern LLM applications increasingly require long context windows, especially for agentic and long-reasoning use cases~\citep{zhuang2025agenticlu,wang2025loongrl}. This imposes three major efficiency challenges on the underlying serving infrastructure: 1) high compute demand for attention in the prefill phase; 2) high memory bandwidth pressure for attention in the decode phase; and 3) high KV cache capacity pressure in the decode phase. Sparse attention~\citep{yuan2025nsa,zhao2025infllmv2,lu2025moba} is a promising technique that has been adopted by recent frontier models~\citep{deepseekai2025deepseekv32,glm5team2026glm5,deepseekai2026deepseekv4} to improve long-context LLM inference efficiency. However, most sparse attention methods primarily reduce attention compute and memory bandwidth; while some also shrink the KV cache, capacity pressure persists as context lengths continue to grow.

A straightforward solution to this problem is to offload the KV cache to CPU memory and only fetch selected KV blocks to GPU memory for sparse attention dynamically during the decode phase. However, CPU fetching through the PCIe interface is still much slower than fetching from GPU memory, which becomes a bottleneck for long-context decoding. Prior work~\citep{zhou2025sparseserve,huang2025nosa} proposes both system- and algorithm-level optimizations to reduce CPU fetching traffic. InfiniGen~\citep{lee2024infinigen} further shows that lookahead prefetching can hide part of this latency, but it relies on the raw hidden state as a proxy for future attention and can be inaccurate when adjacent-layer similarity breaks down.

Meanwhile, sparse attention itself introduces additional overhead from sparse selection. During prefill, a fixed sparse-attention budget reduces attention computation from $O(T^2)$ to $O(T)$, where $T$ is the sequence length, but sparse selection remains $O(T^2)$ and can dominate the attention module as context grows. IndexCache~\citep{bai2026indexcache} and HISA~\citep{xu2026hisa} reduce this overhead through cross-layer index reuse and hierarchical filtering, respectively.

To address the above issues, we propose SparDA, a decoupled sparse attention architecture that introduces a fourth per-layer projection, the \emph{Forecast}, alongside the standard Query, Key, and Value (Figure~\ref{fig:concept}). Rather than treating the previous hidden state as a proxy, SparDA trains this Forecast to estimate next-layer KV importance. SparDA has three key innovations:
\begin{itemize}
\item \textbf{Trainable lookahead sparse selection.} In standard sparse attention, the attention query drives both top-$k$ selection and sparse attention within the same layer. SparDA decouples these roles: the Forecast generated in layer $l$ drives top-$k$ selection for layer $l{+}1$, while the attention query still performs sparse attention, enabling CPU-to-GPU KV cache prefetching to overlap with the current layer's execution.
\item \textbf{Compact Forecast indexer.} Once Forecast is decoupled from the attention query, it no longer needs to preserve the full query-head structure for sparse selection. In our GQA implementation, the Forecast indexer uses a single Forecast head per GQA group, substantially reducing sparse-selection overhead and skipping softmax during selection at inference.
\item \textbf{Asynchronous prefetch with persistent UVA kernel.} We implement the asynchronous prefetch pipeline with a persistent Unified Virtual Addressing (UVA) Triton kernel that performs high-throughput host-to-device transfers in parallel with the main GPU kernels, making sparse-attention offloading practical for large-batch long-context decoding.
\end{itemize}

SparDA can be integrated into sparse-pretrained models by training only the Forecast projections ($<$0.5\% additional parameters on our 8B models) via Kullback--Leibler (KL) divergence against training targets derived from the frozen backbone, without retraining the base model. We evaluate SparDA on MiniCPM4.1-8B~\citep{minicpm2025minicpm4} and NOSA-8B~\citep{huang2025nosa}, two sparse-pretrained 8B models, and find accuracy comparable to each model's full-query sparse-attention baseline on the evaluated long-context and long-reasoning benchmarks. In efficiency, SparDA achieves up to 1.25$\times$ prefill speedup and 1.7$\times$ decode speedup over the sparse-attention offload baseline. By enabling larger feasible batch sizes on a single GPU, SparDA reaches up to 5.3$\times$ higher decode throughput than the non-offload sparse baseline on MiniCPM4.1-8B. We further demonstrate SparDA's architectural transfer to GLM-5.1, a 754B mixture-of-experts model with native token-level DSA~\citep{zai2026glm51}, achieving accuracy comparable to its native selector.

\section{Related Work}
\label{sec:related-work}

\paragraph{KV cache compression and training-free sparse attention.}
KV cache compression discards cached entries to reduce the cache footprint~\citep{zhang2023h2o,li2024snapkv,xiao2025duoattention,xiao2024streamingllm}, which can cause accuracy loss when discarded context is later needed.
QUEST and SparQ select tokens or blocks per query while retaining the full KV cache~\citep{tang2024quest,ribar2024sparq}; RocketKV combines permanent eviction with query-aware sparse attention~\citep{behnam2025rocketkv}. MInference~\citep{jiang2024minference}, FlexPrefill~\citep{lai2025flexprefill}, and XAttention~\citep{xu2025xattention} exploit dynamic sparse patterns to accelerate prefill. Methods retaining the full cache still face growing capacity demands.

\paragraph{Trainable sparse attention.}
Trainable sparse attention learns sparse behavior during pretraining or adaptation. InfLLM-V2~\citep{zhao2025infllmv2} and MoBA~\citep{lu2025moba} use parameter-free block selection over mean-pooled keys; SeerAttention~\citep{gao2025seerattention} learns a gate over pooled queries and keys through self-distillation. Native Sparse Attention (NSA)~\citep{yuan2025nsa} fuses compressed, sparse, and sliding-window branches via learnable gating. DeepSeek Sparse Attention (DSA)~\citep{deepseekai2025deepseekv32} moves to token-level sparsity. DeepSeek-V4~\citep{deepseekai2026deepseekv4} further interleaves Compressed Sparse Attention (CSA) with Heavily Compressed Attention (HCA), cutting per-token KV cache $\sim 10\times$ over DeepSeek-V3.2. However, its $8\times$ longer context still leaves absolute capacity a bottleneck. For reducing sparse selection overhead, IndexCache~\citep{bai2026indexcache} reuses top-$k$ indices across adjacent layers, and HISA~\citep{xu2026hisa} replaces the flat scan with hierarchical filtering. For block-sparse GQA, \method\ adapts DSA's decoupled-indexer design to one-layer-ahead selection with one Forecast head per GQA group.

\paragraph{Sparse attention with KV cache offloading.}
ArkVale~\citep{chen2024arkvale} recalls evicted KV entries from CPU, while ShadowKV~\citep{sun2025shadowkv} combines low-rank key reconstruction with offloaded values. MagicPIG~\citep{chen2025magicpig} uses locality-sensitive hashing to sample entries and computes their attention on CPU. NOSA~\citep{huang2025nosa} pairs a query-agnostic eviction head with a query-aware selector and issues UVA-based CPU-to-GPU transfers synchronously per layer, reducing volume rather than hiding latency. SparseServe~\citep{zhou2025sparseserve} and HiSparse~\citep{lmsys2026sglang_hisparse} offload KV caches to host memory. SparseServe uses working-set-aware batch sizing and fragmentation-aware transfers, while HiSparse combines LRU eviction with a fused fetch kernel. For models that share selections across layers, HiSparse also overlaps future-layer KV prefetch with earlier-layer computation. InfiniGen~\citep{lee2024infinigen} is closest to \method. It uses the raw hidden state as a cross-layer proxy to prefetch a subset of CPU-resident KV to GPU before each layer. These DMA-based memory copies overlap with the preceding layer's execution. \method\ instead learns a cross-layer Forecast projection without requiring shared selections, bridging DSA's trained decoupled indexer with InfiniGen's lookahead. We co-design this projection with a persistent UVA prefetch kernel for communication-computation overlap.

\section{Preliminary}
\label{sec:preliminary}
\subsection{InfLLM-V2}

Our block-sparse implementation builds SparDA on top of InfLLM-V2~\citep{zhao2025infllmv2}. InfLLM-V2 decomposes attention into initial tokens, a local sliding window, and top-$k$ selected blocks. SparDA modifies the top-$k$ selection path while preserving the initial and local components. This section recaps the block-sparse components that SparDA directly extends; the native token-level DSA adaptation is described in Section~\ref{sec:glm_transfer}.

\paragraph{Unified sparse attention.}
InfLLM-V2 merges selected attention and sliding-window attention into a single sparse attention module with shared KV projections $\mathbf{W}_K$ and $\mathbf{W}_V$ inherited from the pretrained dense model. At layer $l$, the attended block set for a query token at position $i$ is:
\begin{equation}
\mathcal{B}_l(i) = \mathcal{B}_{\mathrm{init}} \cup \mathcal{B}_{\mathrm{local}}(i) \cup \mathcal{B}_{\mathrm{topk}}(i),
\end{equation}
where $\mathcal{B}_{\mathrm{init}}$ covers a fixed set of initial blocks, $\mathcal{B}_{\mathrm{local}}(i)$ covers the local blocks, and $\mathcal{B}_{\mathrm{topk}}(i)$ contains the top-$k$ blocks selected by compression scores. Full attention is computed only over the KV entries in $\mathcal{B}_l(i)$, reducing the per-query attention cost from $O(T)$ to $O(|\mathcal{B}_l(i)| \cdot B)$, where $T$ is the sequence length and $B$ is the block size.

\paragraph{Three-stage block representation.}
InfLLM-V2 uses a parameter-free, coarse-to-fine compression pipeline to compute block-level relevance scores for $\mathcal{B}_{\mathrm{topk}}$. In the first stage, keys $\mathbf{K}_l$ are mean-pooled into overlapping compressed representations $\widetilde{\mathbf{K}}_l$ with kernel size $l_{C_1}$ and stride $s_{C_1}$:
\begin{equation}
\widetilde{\mathbf{K}}_{l,j} = \mathrm{Mean}(\mathbf{K}_{l,\, j \cdot s_{C_1} : j \cdot s_{C_1} + l_{C_1}}).
\end{equation}
In grouped-query attention (GQA)~\citep{ainslie2023gqa}, each group $m$ contains $G$ query heads that share one KV head. For head $h$ in group $m$, scoring against the corresponding compressed keys yields per-compressed-key relevance $\mathbf{S}_{l,m}^{h}$. Scores are then summed within each group to produce a shared importance score $\mathbf{S}_{l,m}^{\mathrm{shared}}$.
In the final stage, a max-pooling operation over each $\mathbf{S}_{l,m}^{\mathrm{shared}}$ produces the block-level score used for top-$k$ selection. This three-stage process computes block scores from several overlapping sub-blocks rather than a single coarse pooling, preserving finer-grained information.

\section{Method}
\label{sec:method}

\subsection{SparDA Architecture}

\begin{figure}[t]
\centering
\includegraphics[width=0.99\columnwidth]{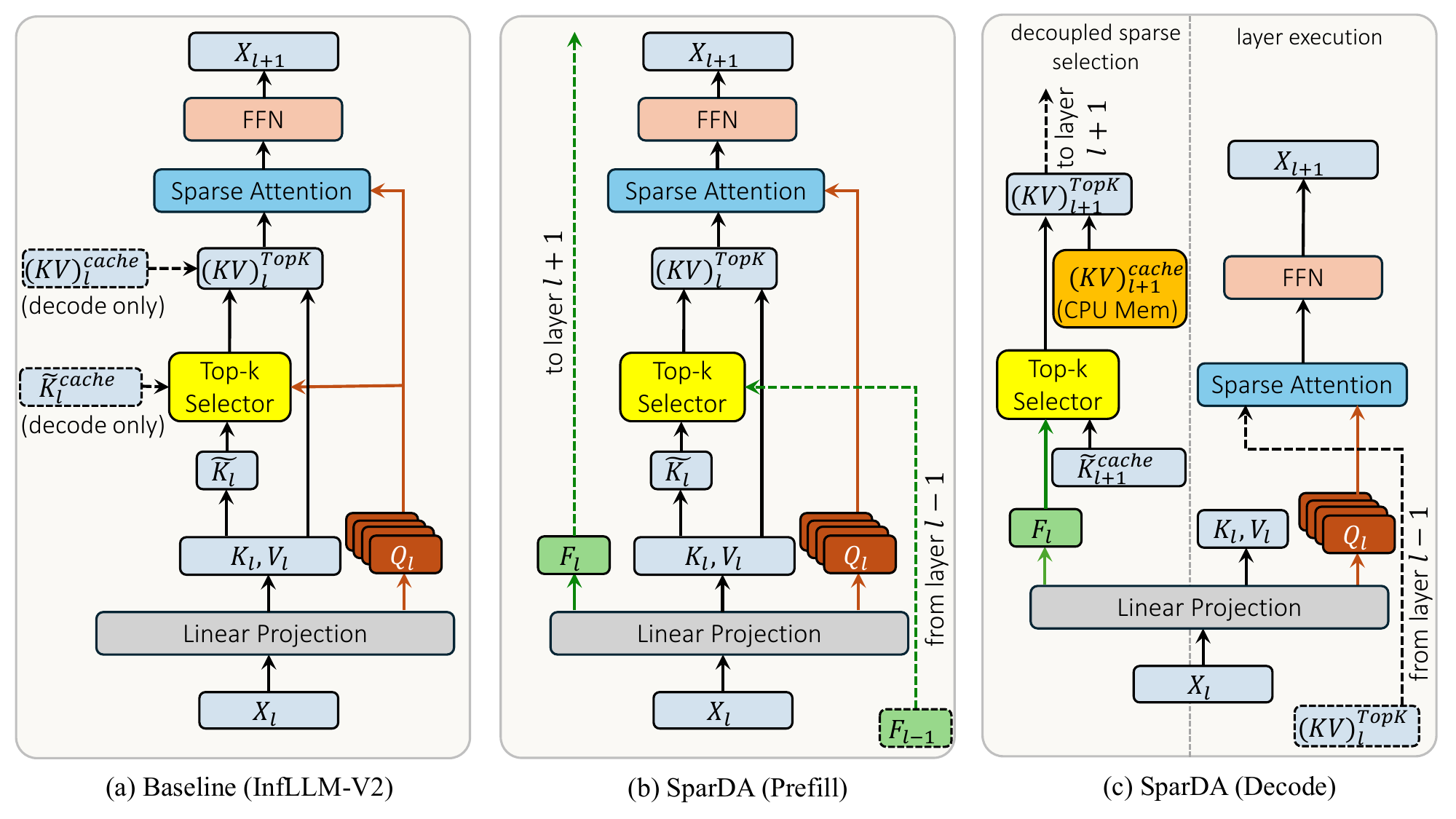}
\caption{\method\ architecture. \textbf{(a)}~In the baseline (InfLLM-V2), $\mathbf{Q}_l$ drives both top-$k$ selection and sparse attention within the same layer, so selection sits on the attention critical path. \textbf{(b, c)}~\method\ adds a Forecast $\mathbf{F}_l$ to the linear projection and uses $\mathbf{F}_{l-1}$ from the previous layer to select blocks for the current layer. This decoupling pays off in two regimes: during prefill~\textbf{(b)} the Forecast replaces the expensive multi-head selector with a Forecast indexer that uses one Forecast head per GQA group; during decode~\textbf{(c)} $\mathbf{F}_l$ predicts $\mathcal{B}_{l+1}$ early enough to prefetch the selected KV blocks from CPU asynchronously, hiding the PCIe transfer behind the current layer's compute.}
\label{fig:architecture_overview}
\end{figure}

As discussed in Section~\ref{sec:introduction}, the primary goals of SparDA are twofold: (1) to enable KV cache prefetching that overlaps with layer execution for latency hiding; and (2) to reduce the overhead of sparse selection. SparDA achieves both through a single architectural change: a fourth per-layer projection, the \emph{Forecast} $\mathbf{F}_l$, produced alongside the standard $\mathbf{Q}_l, \mathbf{K}_l, \mathbf{V}_l$, that drives a compact sparse-selection indexer. $\mathbf{F}_l$ decouples sparse selection from the attention query so it can be computed one layer ahead. This decoupling also means the indexer no longer needs the full query-head structure used for attention. In our GQA implementation, $\mathbf{F}_l$ replaces the original per-query-head scoring loop with one Forecast head per GQA group and skips the softmax normalization. Figure~\ref{fig:architecture_overview} illustrates the resulting layer structure compared with the InfLLM-V2 baseline. The formulation below describes this block-sparse GQA implementation; Section~\ref{sec:glm_transfer} and Appendix~\ref{sec:appendix_glm_transfer} describe the adaptation to native token-level DSA.

\paragraph{Decoupled sparse selection.}
In the original InfLLM-V2 sparse attention (Figure~\ref{fig:architecture_overview}a), the query $\mathbf{Q}_l$ is used both for top-$k$ sparse selection and for the sparse attention computation, and both steps happen within the same layer. SparDA decouples these two roles. The linear projection $\phi_l$ in each layer produces an additional Forecast $\mathbf{F}_l$ besides the standard $\mathbf{Q}_l$, $\mathbf{K}_l$, $\mathbf{V}_l$:
\begin{equation}
(\mathbf{Q}_l, \mathbf{K}_l, \mathbf{V}_l, \mathbf{F}_l) = \phi_l(\mathbf{X}_l).
\end{equation}
For brevity, the equations in this subsection drop the per-query-token index used in Section~\ref{sec:preliminary}. The Forecast predicts the next layer's selection for the same token, so lookahead is across layers and leaves causal masking unchanged. $\mathbf{F}_l$ is used for top-$k$ sparse selection while $\mathbf{Q}_l$ is used only for the sparse attention computation. Let $f_{\mathrm{top}}(\cdot,k)$ denote the InfLLM-V2 block-selection operator, which max-pools compressed-key-grid scores to block-level scores and returns the top-$k$ blocks. Specifically, $\mathbf{F}_l$ scores against the compressed keys $\widetilde{\mathbf{K}}_{l+1}$ of layer $l{+}1$, and the resulting top-$k$ blocks are merged with the initial and local blocks to form the attended set for layer $l{+}1$:
\begin{equation}
\mathcal{B}_{l+1} = \mathcal{B}_{\mathrm{init}} \cup \mathcal{B}_{\mathrm{local}} \cup f_{\mathrm{top}}\!\left( \mathbf{F}_l \widetilde{\mathbf{K}}_{l+1}^{\top},\; k \right).
\end{equation}
For NOSA, the attended set additionally includes blocks chosen by its unchanged query-agnostic selector (Appendix~\ref{sec:appendix_eval}). The sparse attention at layer $l{+}1$ still uses the original query $\mathbf{Q}_{l+1}$:
\begin{equation}
\mathbf{O}_{l+1} =
\mathrm{Attn}\!\left(
\mathbf{Q}_{l+1},\;
\mathbf{K}_{l+1}[\mathcal{B}_{l+1}],\;
\mathbf{V}_{l+1}[\mathcal{B}_{l+1}]
\right).
\end{equation}
For the first layer ($l{=}0$), where no previous Forecast exists, SparDA produces a current-layer Forecast $\mathbf{F}_0^{\mathrm{cur}}$ from $\mathbf{X}_0$ via a separate projection and uses it for same-layer selection. $\mathbf{F}_0^{\mathrm{cur}}$ still uses one Forecast head per GQA group rather than the original attention query, so the selection cost remains lower than the baseline's even without the one-layer-ahead benefit. For the final layer, $\mathbf{F}_{L-1}$ is unused since there is no subsequent layer to predict for.

The benefit of decoupled selection differs across inference phases. During prefill (Figure~\ref{fig:architecture_overview}b), all keys are already on GPU, so the only saving is the reduced sparse selection cost. During decode (Figure~\ref{fig:architecture_overview}c), the KV cache is offloaded to CPU memory. $\mathbf{F}_l$ produces $\mathcal{B}_{l+1}$ while layer $l$ is still executing. This lets the runtime prefetch the selected KV entries from CPU and overlap the PCIe transfer with layer execution. The compressed keys $\widetilde{\mathbf{K}}_{l+1}^{\mathrm{cache}}$ are kept on GPU and updated incrementally. Initial and local blocks are always attended and GPU-resident for both Sparse and \method, and are excluded from the predicted top-$k$ budget. Newly appended keys lie in the local window, so nonlocal block selection does not require the next layer's current-token key. Algorithms~\ref{alg:prefill} and \ref{alg:decode} in Appendix~\ref{sec:appendix_algorithms} summarize the per-layer steps.

\paragraph{Forecast indexer.}
The key observation is that once Forecast is decoupled from the attention query, sparse selection no longer has to use the same head layout as attention. Block-sparse selectors such as InfLLM-V2 currently score each GQA group with all $G$ query heads, because the attention query itself drives selection. DSA's lightning indexer~\citep{deepseekai2025deepseekv32} already exploits this decoupling at the token level with fewer heads than the attention query. SparDA brings the same design to block-sparse attention. $\mathbf{F}_l$ has one Forecast head per GQA group (one per KV head) rather than one per query head. This eliminates the per-query-head scoring loop and significantly reduces top-$k$ selection overhead. It also naturally skips the softmax operation that standard block-sparse selectors apply before top-$k$ ranking, since there is no need for score summation across multiple query heads within a GQA group.

\subsection{Indexer Training}
\label{sec:indexer_training}

SparDA can be added to existing sparse-pretrained models by training only the Forecast projections, without retraining the base model. On both MiniCPM4.1-8B and NOSA-8B, these projections add only 33.5M parameters (0.41\% of the 8B total), making SparDA a lightweight add-on with negligible model-size overhead.

\paragraph{Learning next-layer importance.}
The Forecast learns to estimate which KV entries will matter to the next layer. During training, the frozen model provides that layer's actual attention queries and keys, from which we construct a relevance target. The Forecast from the preceding layer learns to predict this target using the coarse compressed keys available at inference. For the two block-sparse models, we construct the target with finer key compression, so that small groups of relevant tokens are less diluted when forming the supervision signal. The target therefore differs from the Sparse baseline's own coarse selection scores. Only the Forecast projections are optimized.

\paragraph{Aligning the target and prediction.}
The finer target initially has more score positions than the Forecast prediction. We max-pool the target scores onto the prediction's compressed-key grid and normalize both to form comparable distributions. A separate pooling step converts the aligned target scores into block scores for top-$k$ selection. Thus, target blocks are selected \emph{after} pooling to the inference block size, while the loss retains the individual compressed-key scores within those blocks. This separates the choice of important blocks from the finer relevance information used to train the Forecast.

\paragraph{Training objective.}
Inspired by DSA~\citep{deepseekai2025deepseekv32}, we minimize KL divergence from the target distribution to the Forecast prediction. For both distributions, we keep the compressed-key positions within the target-selected blocks individually and combine all remaining probability mass into one ``rest'' entry. Let $\bar p^{\mathrm{tgt}}$ and $\bar p^{\mathrm{pred}}$ denote these normalized distributions. The objective is
\begin{equation}
\mathcal{L}_{\mathrm{KL}} = \mathbb{E}_{l,i,m}\!\left[
\mathrm{KL}\!\left(\bar p^{\mathrm{tgt}}_{l,i,m}\,\|\,\bar p^{\mathrm{pred}}_{l,i,m}\right)
\right],
\end{equation}
where the average is over layers $l$, query tokens $i$, and GQA groups $m$. The loss teaches the Forecast both the relative importance of selected regions and how much probability belongs elsewhere. Softmax is used to form distributions during training; inference only needs score rankings. Appendix~\ref{sec:appendix_training} gives the pooling settings and training details, and Table~\ref{tab:ablation_accuracy} compares the target resolutions. The GLM adaptation uses its native DSA indexer as the teacher, as described in Section~\ref{sec:glm_transfer}.

\subsection{Efficient Implementation}
\label{sec:efficient_impl}

At inference, SparDA's one-layer-ahead Forecast produces $\mathcal{B}_{l+1}$ while layer $l$ is still executing. SparDA exploits this early availability through an asynchronous prefetch pipeline that transfers only the selected KV blocks from CPU to GPU before they are needed.

\paragraph{Persistent UVA kernel for asynchronous prefetch.}
Once the sparse pattern for the next layer is predicted, the runtime reuses GPU-resident blocks in $\mathcal{B}_{l+1}$ and fetches only the missing blocks from pinned CPU memory on a dedicated CUDA stream so that the transfer overlaps with layer execution. Instead of relying on many small, irregular memory copies, SparDA uses a persistent Triton~\citep{tillet2019triton} kernel based on Unified Virtual Addressing (UVA). The kernel keeps a small fixed set of GPU thread blocks, or Cooperative Thread Arrays (CTAs), active and lets them continuously process block-transfer tasks within a single launch, reducing launch overhead, avoiding frequent synchronization, and limiting interference with the main compute stream. As a result, CPU-to-GPU KV cache transfers can be largely overlapped with layer execution, making prefetching practical even with large-batch, high-throughput decoding.

\paragraph{Batch-adaptive CTA allocation.}
The number of CTAs in the persistent kernel controls the trade-off between prefetch throughput and layer execution speed. More CTAs accelerate transfers toward the PCIe bandwidth ceiling but consume Streaming Multiprocessors (SMs) that would otherwise run the attention and FFN kernels. At small batch sizes the GPU is underutilized and layer execution dominates the critical path; a modest CTA budget suffices because the total transfer volume is small. As batch size grows, layer execution time scales sub-linearly (better SM occupancy) while prefetch volume grows linearly. Beyond a crossover point the prefetch becomes the new bottleneck, and allocating additional CTAs is worthwhile despite the marginal slowdown in layer execution. SparDA therefore adopts a simple heuristic to choose the CTA count that maximizes overall throughput for the underlying hardware. The sweep and chosen configurations are reported in Table~\ref{tab:prefetch_cta_sweep} (Appendix~\ref{sec:appendix_additional}).

\section{Experiments}
\label{sec:experiments}

\subsection{Experimental Setup}

Our main accuracy and efficiency evaluation uses two 8B models: MiniCPM4.1-8B~\citep{minicpm2025minicpm4}, whose sparse backbone is InfLLM-V2~\citep{zhao2025infllmv2}, and NOSA-8B~\citep{huang2025nosa}, which adds a query-agnostic eviction head on top of InfLLM-V2. Accuracy is measured on HELMET~\citep{yen2025helmet}, LongBench~\citep{bai2024longbench}, RULER~\citep{hsieh2024ruler}, and a long-reasoning suite (MATH-500~\citep{lightman2023letsverify}, AIME 2024~\citep{zhang2024aime}, AIME 2025~\citep{zhang2025aime}). We compare four configurations on these models: dense, sparse, InfiniGen~\citep{lee2024infinigen}, and \method. We additionally evaluate architectural transfer on GLM-5.1~\citep{zai2026glm51}, comparing accuracy against its native token-level DSA selector. We do not directly compare against DSA-specific accelerators such as IndexCache~\citep{bai2026indexcache} or HISA~\citep{xu2026hisa}. Efficiency is measured for the 8B models on NVIDIA H100 and A100 GPUs using the runtime described in Section~\ref{sec:efficient_impl}. Dense$^\dagger$ and Sparse$^\dagger$ denote the no-offload configurations. Sparse, InfiniGen, and \method\ use CPU offloading. The main text reports H100 results; Appendix~\ref{sec:appendix_additional} includes A100 results. Evaluation details for the 8B models and GLM-5.1 are in Appendices~\ref{sec:appendix_eval} and~\ref{sec:appendix_glm_transfer}, respectively.

\subsection{Accuracy Results}

\paragraph{Aggregate accuracy.}
Across the four benchmark families, the results support accuracy comparable to the full-query Sparse baseline (Table~\ref{tab:main_accuracy}). On MiniCPM4.1-8B, \method\ reaches an average of 61.9 versus 61.7 for Sparse, with reasoning scores of 85.7 versus 84.5, a higher RULER score (+0.5), LongBench essentially flat, and HELMET slightly lower (-0.6). On NOSA-8B, the averages are 51.1 versus 49.3, with reasoning scores of 54.8 versus 50.5, higher RULER (+1.7) and HELMET (+1.2) scores, and slightly lower LongBench (-0.1). All four configurations use the eight-seed AIME means reported in Appendix~\ref{sec:appendix_eval} (Table~\ref{tab:aime_multiseed}) for their reasoning averages. \method's positive AIME mean differences over Sparse are within the reported seed-to-seed variation, and we do not claim statistically significant gains from these runs. Across benchmark families and individual tasks, the observed differences can be positive or negative. Dense scores remain higher than both Sparse and \method\ on HELMET and RULER at the models' native maximum sequence lengths (64K for MiniCPM4.1-8B, 32K for NOSA-8B). InfiniGen also has lower HELMET and RULER scores than Sparse on both models. LongBench differences are mixed. Its reasoning average is tied on MiniCPM at the displayed precision and higher on NOSA. Per-task breakdowns are in Appendix~\ref{sec:appendix_detailed_accuracy}.

\begin{table*}[!htbp]
\centering
\small
\setlength{\tabcolsep}{4pt}
\caption{Aggregated benchmark averages. Reasoning averages the single-run MATH-500 score and eight-seed AIME 2024/2025 means for every method (Table~\ref{tab:aime_multiseed}). \textit{Avg} is the arithmetic mean of the four benchmark families, computed before rounding. Best values among sparse methods (excluding Dense) are \textbf{bolded}.}
\label{tab:main_accuracy}
\resizebox{\textwidth}{!}{%
\begin{tabular}{l|rrrrr|rrrrr}
\toprule
 & \multicolumn{5}{c|}{MiniCPM4.1-8B} & \multicolumn{5}{c}{NOSA-8B} \\
Method & HELMET & LongBench & RULER & Reasoning & Avg & HELMET & LongBench & RULER & Reasoning & Avg \\
\midrule
Dense & 41.7 & 44.8 & 85.3 & 85.9 & 64.4 & 39.3 & 42.5 & 86.2 & 52.0 & 55.0 \\
Sparse & \textbf{38.9} & 45.0 & 78.2 & 84.5 & 61.7 & 32.2 & \textbf{42.4} & 72.2 & 50.5 & 49.3 \\
InfiniGen & 33.5 & \textbf{45.1} & 68.4 & 84.5 & 57.9 & 28.1 & 41.6 & 65.2 & 54.4 & 47.3 \\
SparDA & 38.3 & \textbf{45.1} & \textbf{78.7} & \textbf{85.7} & \textbf{61.9} & \textbf{33.4} & 42.3 & \textbf{73.9} & \textbf{54.8} & \textbf{51.1} \\
\bottomrule
\end{tabular}
}
\end{table*}

\paragraph{Length generalization.}
\method\ has higher measured RULER scores than Sparse at each evaluated sequence length on both models (Table~\ref{tab:ablation_ruler_seqlen}). On NOSA-8B, the observed gap widens steadily from +1.7 at 32K to +4.3 at 128K; MiniCPM4.1-8B shows a more mixed pattern (+1.5 at 32K, +0.5 at 64K, +2.1 at 96K, +1.1 at 128K). These measurements support using the Forecast at the evaluated extended contexts; they do not establish a general accuracy advantage across tasks.

\begin{table*}[!htbp]
\centering
\small
\setlength{\tabcolsep}{6pt}
\caption{RULER average accuracy at extended sequence lengths. \method\ has higher measured scores at all evaluated lengths on both models; the observed gap widens steadily with sequence length on NOSA-8B.}
\label{tab:ablation_ruler_seqlen}
\begin{tabular}{l|rrrr|rrrr}
\toprule
 & \multicolumn{4}{c|}{MiniCPM4.1-8B} & \multicolumn{4}{c}{NOSA-8B} \\
Method & 32K & 64K & 96K & 128K & 32K & 64K & 96K & 128K \\
\midrule
Sparse & 86.1 & 78.2 & 68.7 & 67.7 & 72.2 & 56.6 & 48.8 & 40.7 \\
SparDA & \textbf{87.6} & \textbf{78.7} & \textbf{70.8} & \textbf{68.8} & \textbf{73.9} & \textbf{60.5} & \textbf{52.9} & \textbf{45.0} \\
$\Delta$ & +1.5 & +0.5 & +2.1 & +1.1 & +1.7 & +3.9 & +4.1 & +4.3 \\
\bottomrule
\end{tabular}
\end{table*}

\FloatBarrier
\begin{samepage}
\paragraph{Transfer to token-level DSA.}
\label{sec:glm_transfer}
To test transfer beyond 8B block-sparse models, we adapt \method\ to GLM-5.1, a 754B mixture-of-experts model with native token-level DSA~\citep{zai2026glm51,glm5team2026glm5}. We freeze the backbone and train the Forecast at 64K context against the native DSA indexer's selection distribution, rather than the full attention query. The Forecast retains the native indexer's 32 heads, each with dimension 128. This indexer is already compact relative to the model's 64 attention heads. Reducing the Forecast head count further degraded accuracy. This adaptation therefore does not provide the prefill selection-compute savings of our GQA implementation. For GLM-5.1, we use the Forecast during decoding and native DSA during prefill. Across RULER, HELMET, LongBench, and reasoning, \method\ obtains an average score of 80.20 versus 80.10 for native DSA (Table~\ref{tab:glm_accuracy}), supporting comparable accuracy at this larger scale and with token-level selection. Appendix~\ref{sec:appendix_glm_transfer} gives the evaluation settings and detailed results. These results support architectural transfer with comparable accuracy.
\par
\end{samepage}

\begin{table*}[!htbp]
\centering
\small
\setlength{\tabcolsep}{6pt}
\caption{GLM-5.1 aggregate accuracy. HELMET, LongBench, and RULER use the 64K setting; reasoning uses a 32K generation budget. Reasoning averages MATH-500 and the eight-seed AIME means. \textit{Avg} averages the four benchmark families. Best values are \textbf{bolded}.}
\label{tab:glm_accuracy}
\begin{tabular}{l|rrrrr}
\toprule
 & \multicolumn{5}{c}{GLM-5.1 (754B MoE)} \\
Method & HELMET & LongBench & RULER & Reasoning & Avg \\
\midrule
Native DSA & \textbf{71.07} & \textbf{55.52} & \textbf{96.46} & 97.37 & 80.10 \\
SparDA & 70.58 & 55.40 & 96.42 & \textbf{98.41} & \textbf{80.20} \\
\bottomrule
\end{tabular}
\end{table*}

\FloatBarrier
\subsection{Efficiency Results}

\begin{figure}[!t]
\centering
\includegraphics[width=\columnwidth]{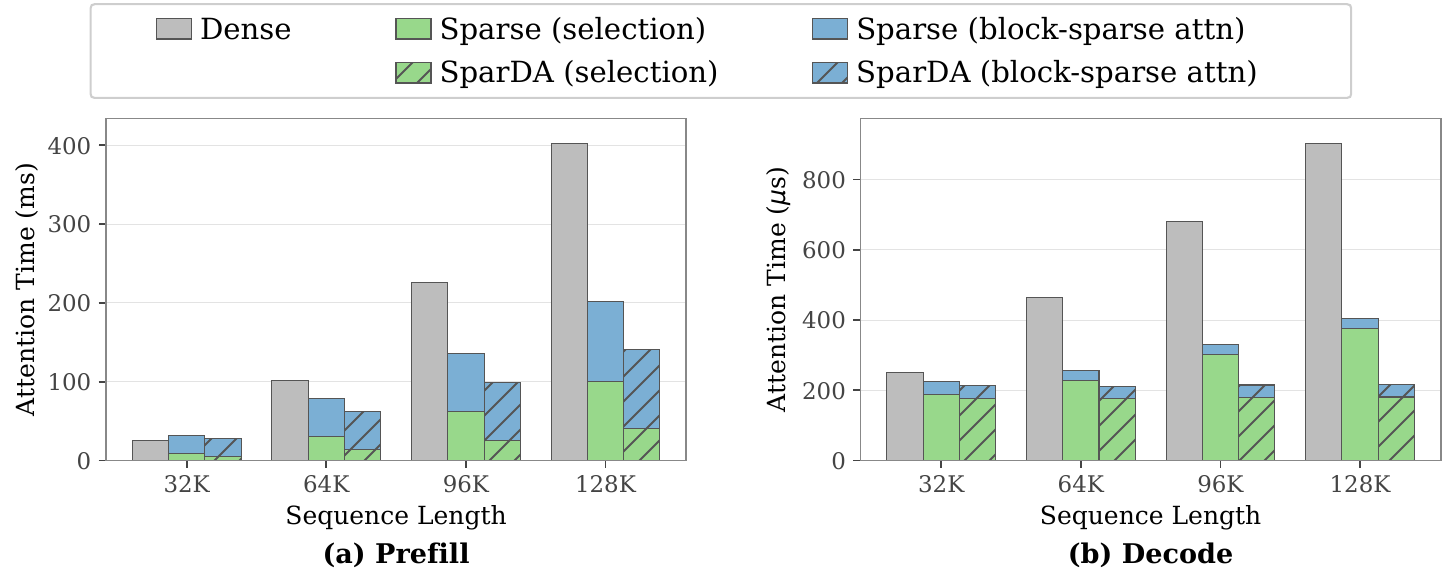}
\caption{Per-layer attention wall time on MiniCPM4.1-8B at batch size 4. \textbf{(a)}~Prefill (ms). \textbf{(b)}~Decode ($\mu$s). Each group contains a Dense bar (gray) and Sparse/SparDA bars (solid/hatched) stacked by block selection (green) and block-sparse attention (blue). SparDA significantly reduces block-selection cost for both prefill and decode.}
\label{fig:attn_breakdown}
\end{figure}

\paragraph{Attention breakdown.}
To understand where the speedup comes from, we break down per-layer attention time on MiniCPM4.1-8B at batch size 4 into block selection (green) and block-sparse attention (blue) for Sparse and \method\ (Figure~\ref{fig:attn_breakdown}). We focus on MiniCPM4.1-8B because NOSA-8B includes additional query-agnostic components that obscure the comparison of query-aware sparse attention, the core component shared by most sparse attention methods. During prefill (Figure~\ref{fig:attn_breakdown}a), both block-sparse attention and block selection grow with sequence length, with selection becoming comparable to attention at 128K. \method\ reduces block-selection cost by approximately 2.5$\times$ at 128K while keeping block-sparse attention time comparable to Sparse. During decode (Figure~\ref{fig:attn_breakdown}b), block-sparse attention is cheap because each step has only one query token. Block selection becomes the bottleneck for Sparse, and its cost grows with sequence length. \method's Forecast indexer keeps decode-time selection nearly flat, cutting the overhead by more than 2$\times$ at 128K. The near-flat decode profile reflects that the indexer's compute is small enough to leave the GPU under-utilized, so longer contexts barely affect decode latency.

\paragraph{Prefill throughput.}
The selection speedup translates directly into prefill throughput gains (Table~\ref{tab:prefill_throughput}). We report a single batch size of 4 because prefill throughput is insensitive to batch size, as GPUs are already fully utilized during the compute-heavy prefill phase. Dense attention leads at short sequences but its quadratic scaling causes it to fall behind as sequence length grows. \method\ consistently outperforms all other methods from 64K onward on MiniCPM4.1-8B, achieving up to 1.25$\times$ speedup over Sparse and 2.11$\times$ over Dense at 128K. On NOSA-8B, \method\ leads from 96K onward with up to 1.16$\times$ over Sparse and 1.40$\times$ over Dense at 128K. All sparse methods are slower on NOSA-8B because its query-agnostic eviction head adds extra work. The main source of \method's speedup is the reduced block-selection cost of its Forecast indexer (Figure~\ref{fig:attn_breakdown}a), which scales more favorably than the baseline's multi-head selector as sequence length increases. KV cache offloading itself has negligible impact on prefill throughput. Sparse$\dagger$ (no offload) and Sparse (with offload) achieve nearly identical throughput. The only offload-time transfer is the asynchronous CPU writeback of newly created KV cache entries after they are used for attention.

\begin{table*}[!htbp]
\centering
\small
\caption{Prefill throughput (tok/s) on NVIDIA H100 GPU at batch size 4. Sequence length swept from 32K to 128K. $\dagger$: no offload. Best value per column is \textbf{bolded}.}
\label{tab:prefill_throughput}
\begin{tabular}{l|rrrr|rrrr}
\toprule
 & \multicolumn{4}{c|}{MiniCPM4.1-8B} & \multicolumn{4}{c}{NOSA-8B} \\
Method & 32K & 64K & 96K & 128K & 32K & 64K & 96K & 128K \\
\midrule
Dense$\dagger$ & \textbf{20388.3} & 13673.7 & 10228.3 & 8085.8 & \textbf{20438.9} & \textbf{13701.0} & 10244.1 & 8118.0 \\
Sparse$\dagger$ & 18706.6 & 16377.3 & 14774.1 & 13676.1 & 12896.5 & 11418.0 & 10514.7 & 9832.5 \\
Sparse & 18548.4 & 16254.4 & 14707.7 & 13661.8 & 12778.3 & 11359.2 & 10448.5 & 9805.2 \\
InfiniGen & 18445.3 & 16249.9 & 14749.2 & 13643.9 & 12838.5 & 11372.4 & 10501.6 & 9749.3 \\
SparDA & 19845.6 & \textbf{18379.5} & \textbf{17715.2} & \textbf{17087.6} & 13456.0 & 12386.1 & \textbf{11807.3} & \textbf{11332.7} \\
\bottomrule
\end{tabular}
\end{table*}

\paragraph{Decode throughput.}
Decode is where the lookahead design pays off most, as it lets CPU-to-GPU KV prefetch overlap with the current layer's compute. Within the offload regime at 128K, \method\ achieves up to 1.69$\times$ speedup over Sparse on MiniCPM4.1-8B and 1.40$\times$ on NOSA-8B (Table~\ref{tab:throughput}), driven by reduced block-selection overhead (Figure~\ref{fig:attn_breakdown}b) and overlapped prefetch that hides the otherwise synchronous CPU-to-GPU transfer. The largest speedups appear at middle batch sizes (B8--B16), where prefetch and layer execution are roughly balanced. Against the non-offload baselines ($\dagger$), which OOM past B16 at long contexts, offloading lets \method\ run much larger feasible batches and obtain correspondingly larger throughput gains, up to $5.28\times$ over Sparse$\dagger$ and $9.21\times$ over Dense$\dagger$ on MiniCPM4.1-8B. Against the offload baseline, NOSA-8B shows lower speedups because its query-agnostic eviction head already reduces KV fetch traffic, leaving less room for overlap. InfiniGen is consistently slower than Sparse primarily because it gathers top-$k$ KV blocks on CPU before transferring them to GPU, making CPU-side gather the bottleneck. Because \method\ retains the layer-0 KV cache on GPU (Appendix~\ref{sec:appendix_eval}), it OOMs slightly earlier than Sparse at long contexts. Nevertheless, \method's peak throughput exceeds Sparse's at every context length, even when Sparse runs at a larger batch.

\begin{table*}[!htbp]
\centering
\small
\setlength{\tabcolsep}{3pt}
\caption{Decode throughput (tok/s) on NVIDIA H100 GPU. Batch size (B) swept from 4 to 128. $\dagger$: no offload. ``--'': OOM. Best value per column is \textbf{bolded}.}
\label{tab:throughput}
\resizebox{\textwidth}{!}{%
\begin{tabular}{cl|rrrrrr|rrrrrr}
\toprule
& & \multicolumn{6}{c|}{MiniCPM4.1-8B} & \multicolumn{6}{c}{NOSA-8B} \\
Context & Method & B4 & B8 & B16 & B32 & B64 & B128 & B4 & B8 & B16 & B32 & B64 & B128 \\
\midrule
 & Dense$\dagger$ & 248.3 & 352.2 & 440.2 & -- & -- & -- & \textbf{252.0} & 350.7 & 439.7 & -- & -- & -- \\
 & Sparse$\dagger$ & \textbf{277.2} & \textbf{550.9} & \textbf{1073.0} & -- & -- & -- & 241.3 & \textbf{488.1} & \textbf{947.1} & -- & -- & -- \\
32K & Sparse & 224.4 & 420.9 & 661.5 & 861.8 & 1088.4 & 1225.1 & 244.0 & 478.0 & 795.8 & 1244.9 & 1861.6 & 2145.5 \\
 & InfiniGen & 56.3 & 75.1 & 98.8 & 140.4 & 184.8 & 222.4 & 82.0 & 120.3 & 166.0 & 205.9 & 269.9 & 356.5 \\
 & SparDA & 235.7 & 471.9 & 918.5 & \textbf{1205.2} & \textbf{1328.0} & \textbf{1358.2} & 214.7 & 433.2 & 873.2 & \textbf{1579.9} & \textbf{2106.6} & \textbf{2312.1} \\
\midrule
 & Dense$\dagger$ & 174.1 & 218.3 & -- & -- & -- & -- & 174.3 & 217.6 & -- & -- & -- & -- \\
 & Sparse$\dagger$ & \textbf{246.5} & 466.2 & -- & -- & -- & -- & \textbf{233.2} & \textbf{439.7} & -- & -- & -- & -- \\
64K & Sparse & 213.1 & 350.8 & 549.4 & 756.1 & 937.2 & 1041.9 & 221.0 & 397.9 & 704.6 & 1071.1 & 1553.1 & 1793.1 \\
 & InfiniGen & 55.2 & 63.3 & 94.0 & 129.1 & 166.5 & -- & 78.1 & 106.2 & 139.1 & 181.0 & 220.4 & -- \\
 & SparDA & 234.9 & \textbf{470.3} & \textbf{811.8} & \textbf{1026.5} & \textbf{1049.9} & \textbf{1159.1} & 218.8 & 431.2 & \textbf{870.9} & \textbf{1392.3} & \textbf{1905.2} & \textbf{1996.3} \\
\midrule
 & Dense$\dagger$ & 133.8 & 159.0 & -- & -- & -- & -- & 130.0 & 158.3 & -- & -- & -- & -- \\
 & Sparse$\dagger$ & 215.5 & 401.4 & -- & -- & -- & -- & 204.1 & 380.1 & -- & -- & -- & -- \\
96K & Sparse & 185.7 & 306.0 & 489.0 & 679.4 & 839.5 & \textbf{934.3} & 191.5 & 341.3 & 605.5 & 962.4 & 1354.3 & \textbf{1580.5} \\
 & InfiniGen & 49.4 & 64.9 & 86.0 & 117.9 & -- & -- & 75.1 & 104.0 & 129.1 & 165.2 & -- & -- \\
 & SparDA & \textbf{238.6} & \textbf{475.6} & \textbf{752.9} & \textbf{939.2} & \textbf{1076.3} & -- & \textbf{215.5} & \textbf{437.6} & \textbf{852.9} & \textbf{1263.7} & \textbf{1694.1} & -- \\
\midrule
 & Dense$\dagger$ & 108.6 & -- & -- & -- & -- & -- & 108.4 & -- & -- & -- & -- & -- \\
 & Sparse$\dagger$ & 189.5 & -- & -- & -- & -- & -- & 179.3 & -- & -- & -- & -- & -- \\
128K & Sparse & 167.8 & 279.5 & 447.9 & 618.6 & 788.9 & -- & 173.2 & 285.4 & 529.2 & 898.7 & 1298.3 & -- \\
 & InfiniGen & 51.8 & 66.5 & 85.6 & 117.5 & -- & -- & 77.5 & 105.2 & 131.7 & 166.9 & -- & -- \\
 & SparDA & \textbf{240.2} & \textbf{471.2} & \textbf{705.3} & \textbf{899.2} & \textbf{1000.1} & -- & \textbf{219.0} & \textbf{399.3} & \textbf{735.0} & \textbf{1127.0} & \textbf{1463.3} & -- \\
\bottomrule
\end{tabular}
}
\end{table*}

Appendix~\ref{sec:appendix_additional} provides a decode speedup breakdown that separates block-selection savings from prefetch overlap, ablations on the compression window and prefetch CTA allocation, and throughput results on NVIDIA A100 GPU.

\section{Limitations}
\label{sec:limitations}

SparDA builds on an existing sparse attention backbone to improve inference efficiency. The decoupled indexer retains the base model and sparse attention computation, but it can change which KV entries are selected. Its accuracy therefore depends on both the base model and the Forecast's selection decisions. Our GLM-5.1 results extend the accuracy evaluation to token-level DSA at 754B scale; efficiency measurements remain limited to the two 8B models. We have not measured SparDA's throughput benefit on GLM-5.1 because integrating KV offloading and lookahead prefetch into its multi-node serving stack requires substantial additional engineering. GLM-5.1 is evaluated in SGLang across four nodes of eight H100 GPUs, with tensor, pipeline, and expert parallelism, whereas our prefetch runtime currently targets one GPU per replica. Coordinating per-rank KV partitions, pipeline-stage boundaries, and expert all-to-all communication makes this integration complex. Measuring the resulting throughput and extending the method to other sparse architectures, including the Compressed Sparse Attention (CSA) path of DeepSeek-V4~\citep{deepseekai2026deepseekv4}, remain future work.

\section{Conclusion}
\label{sec:conclusion}
\method\ shows that sparse selection can be treated as a trainable, schedulable signal rather than an operation tied to the current attention query. By decoupling selection from attention, the Forecast projection predicts block choices one layer ahead, letting the runtime overlap CPU-to-GPU KV transfer with current-layer execution while reducing selector cost through a compact GQA-level indexer. More broadly, the results suggest that sparse attention mechanisms should expose future memory-access patterns early enough for the serving system to act on them, turning sparsity from a compute-saving device into an offloading-friendly schedule. On the two evaluated 8B models, \method\ maintains accuracy comparable to the full-query sparse-attention baseline while improving long-context inference efficiency without retraining the base model. Our adaptation to GLM-5.1 (754B MoE) also achieves accuracy comparable to its native token-level selector. This provides evidence that decoupled selection transfers beyond block-sparse 8B models.

\begin{ack}
This work was funded by NVIDIA. Yaosheng Fu, Song Han, and Oreste Villa are employed by NVIDIA, and Song Han is also affiliated with MIT. Guangxuan Xiao and Xin Dong contributed to this work while employed at NVIDIA and are now with Thinking Machines Lab and ByteDance Seed, respectively. The authors declare no other competing interests.
\end{ack}

\bibliographystyle{plainnat}
\bibliography{refs}

\newpage
\appendix
\section{Algorithm Pseudocode}
\label{sec:appendix_algorithms}

\begin{algorithm}[h]
\small
\DontPrintSemicolon
\caption{\method\ prefill step (all tokens, one layer)}\label{alg:prefill}
\KwIn{Hidden states $\mathbf{X}_l$; Forecast $\mathbf{F}_{l-1}$ from prior layer}
\KwOut{Hidden states $\mathbf{X}_{l+1}$; Forecast $\mathbf{F}_l$ for next layer}
$(\mathbf{Q}_l, \mathbf{K}_l, \mathbf{V}_l, \mathbf{F}_l) \gets \phi_l(\mathbf{X}_l)$ \tcp*{$\mathbf{F}_l$ is the Forecast}
$\mathcal{B}_l \gets \mathcal{B}_{\mathrm{init}} \cup \mathcal{B}_{\mathrm{local}} \cup f_{\mathrm{top}}(\mathbf{F}_{l-1}\, \widetilde{\mathbf{K}}_{l}^{\top},\; k)$ \tcp*{pooled block selection}
$\mathbf{O}_l \gets \mathrm{Attn}(\mathbf{Q}_l,\, \mathbf{K}_l[\mathcal{B}_l],\, \mathbf{V}_l[\mathcal{B}_l])$\;
$\mathbf{X}_{l+1} \gets \mathrm{FFN}_l(\mathbf{O}_l)$\;
\Return $\mathbf{X}_{l+1},\, \mathbf{F}_l$\;
\end{algorithm}

For the first layer ($l{=}0$), where no previous Forecast exists, \method\ uses the layer-0 Forecast $\mathbf{F}_0^{\mathrm{cur}}$ described in Section~\ref{sec:method} for same-layer selection. In decode, the layer-0 KV cache remains on GPU, so the prefetch path begins from subsequent layers. At the final layer ($l{=}L{-}1$), the next-layer selection and prefetch steps in Algorithm~\ref{alg:decode} are skipped.

\begin{algorithm}[h]
\small
\DontPrintSemicolon
\caption{\method\ decode step (one token, one layer)}\label{alg:decode}
\KwIn{Hidden state $\mathbf{X}_l$; selected blocks $\mathcal{B}_l$ and prefetched KV entries from prior layer}
\KwOut{Hidden state $\mathbf{X}_{l+1}$; selected blocks $\mathcal{B}_{l+1}$ for the next layer}
$(\mathbf{Q}_l, \mathbf{K}_l, \mathbf{V}_l, \mathbf{F}_l) \gets \phi_l(\mathbf{X}_l)$ \tcp*{$\mathbf{F}_l$ is the Forecast}
Append $(\mathbf{K}_l, \mathbf{V}_l)$ to $(\mathbf{K}_l^{\mathrm{cache}}, \mathbf{V}_l^{\mathrm{cache}})$\;
Update $\widetilde{\mathbf{K}}_l^{\mathrm{cache}}$ incrementally from $\mathbf{K}_l^{\mathrm{cache}}$\;
$\mathcal{B}_{l+1} \gets \mathcal{B}_{\mathrm{init}} \cup \mathcal{B}_{\mathrm{local}} \cup f_{\mathrm{top}}(\mathbf{F}_l\, \widetilde{\mathbf{K}}_{l+1}^{\mathrm{cache}\top},\; k)$ \tcp*{one-layer-ahead pooled block selection}
Launch prefetch of missing blocks in $\mathcal{B}_{l+1}$: CPU$\to$GPU on prefetch stream\;
Wait for the KV entries of $\mathcal{B}_l$ (if not yet complete) \tcp*{prefetch launched during layer $l{-}1$}
$\mathbf{O}_l \gets \mathrm{Attn}(\mathbf{Q}_l,\, \mathbf{K}_l^{\mathrm{cache}}[\mathcal{B}_l],\, \mathbf{V}_l^{\mathrm{cache}}[\mathcal{B}_l])$\;
$\mathbf{X}_{l+1} \gets \mathrm{FFN}_l(\mathbf{O}_l)$\;
\Return $\mathbf{X}_{l+1},\, \mathcal{B}_{l+1}$\;
\end{algorithm}

\section{Training Details}
\label{sec:appendix_training}

\paragraph{Target construction and alignment.}
For the block-sparse models, mean-pool the frozen backbone's keys with kernel size 2 and stride 1 to form the training target, and with kernel size 32 and stride 16 for the Forecast prediction. The target uses the actual queries of the layer being predicted: apply causal softmax to each query head's scores and sum within each GQA group. Max-pool these fine target scores onto the prediction's $(32,16)$ compressed-key grid, then normalize. The Forecast produces a distribution on that same grid using its own scores and causal softmax. Key compression, alignment of target scores, and pooling scores into final KV blocks are distinct operations.

\paragraph{Block selection and loss.}
To choose the target top-$k$ blocks, max-pool the aligned target scores to the 64-token block granularity. Initial and local blocks are always included in attention and excluded from this top-$k$ choice. For the loss, return to the shared compressed-key grid and retain each score position within the selected blocks. Apply the same selection to target and prediction, combine the remaining mass into one rest entry, and normalize before taking $\mathrm{KL}(\mathrm{target}\,\|\,\mathrm{prediction})$. The loss therefore has one entry per retained compressed-key position plus the rest entry, rather than one entry per selected block. The backbone and target remain fixed; gradients update only the Forecast projections. Layer 0 uses the same-layer Forecast defined in Section~\ref{sec:method}.

Table~\ref{tab:training_hparams} lists the indexer training hyperparameters used for both the MiniCPM4.1-8B and NOSA-8B experiments. Only the Forecast projection weights are trained; all backbone parameters remain frozen. Training uses 32 NVIDIA H100 GPUs; MiniCPM4.1-8B at 64K completes within 48 hours, and NOSA-8B at 32K completes within 24 hours.

\begin{table}[h]
\centering
\small
\caption{Indexer training hyperparameters.}
\label{tab:training_hparams}
\begin{tabular}{ll}
\toprule
Hyperparameter & Value \\
\midrule
Optimizer & AdamW~\citep{loshchilov2019adamw} ($\beta_1{=}0.9$, $\beta_2{=}0.95$, $\epsilon{=}10^{-6}$) \\
Weight decay & 0.01 \\
Learning rate & $5 \times 10^{-4}$ (constant) \\
Training steps & 2\,000 optimizer steps \\
Effective batch size & 32 (accumulated across GPUs) \\
Precision & BF16 mixed precision \\
Gradient clipping & Max norm 0.5 \\
Training data & ProLong-64K \citep{gao2024prolong} \\
Sequence length (MiniCPM4.1-8B) & 65\,536 tokens \\
Sequence length (NOSA-8B) & 32\,768 tokens \\
Compression window & Target $(l_{C_1}{=}2, s_{C_1}{=}1)$; prediction $(l_{C_1}{=}32, s_{C_1}{=}16)$ \\
\bottomrule
\end{tabular}
\end{table}

\section{Evaluation Details}
\label{sec:appendix_eval}

\paragraph{Model configuration.}
Both MiniCPM4.1-8B and NOSA-8B use a block size of 64 tokens, 1 initial block, and a compression kernel of size 32 with stride 16 for mean-pooled keys. Both models use a total attended-block budget of 96, including the forced initial and local blocks. The $k$ in the method equations counts only query-aware predicted blocks, excluding these forced blocks. MiniCPM4.1-8B uses a local window of 2048 tokens (32 blocks) following its default sparse configuration, with the remaining budget after initial/local blocks assigned to query-aware selection. NOSA-8B uses a local window of 1024 tokens (16 blocks) and allocates 24 blocks (25\% of 96) for query-aware selection. Its query-agnostic eviction head fills the remaining budget, following \citet{huang2025nosa}. For MiniCPM4.1-8B, the thinking mode is disabled for all benchmarks except the reasoning suite. For the \method\ indexer, the trained Forecast projections (the main projection and the layer-0 same-layer projection) are loaded from the checkpoint. All other model parameters are identical to the sparse baseline.

\paragraph{Long-context extension.}
MiniCPM4.1-8B natively supports up to 64K tokens and NOSA-8B up to 32K. For benchmarks that exceed each model's native maximum, we apply positional-encoding extension at inference time. For MiniCPM4.1-8B we use the official 128K-validated LongRoPE~\citep{ding2024longrope} frequency factors provided with the model checkpoint, which replace both the long-factor and short-factor tables in the rotary embedding. For NOSA-8B we increase the RoPE base frequency (\texttt{rope\_theta}) from the default 10,000 to 40,000 following \citet{huang2025nosa}, a simple scaling that extends the effective context window to 128K. The same positional extension is applied identically across all attention configurations (Dense, Sparse, InfiniGen, \method) so that any accuracy differences are attributable solely to the attention method, not the positional encoding.

\paragraph{InfiniGen adaptation.}
InfiniGen~\citep{lee2024infinigen} is originally a token-level sparse attention method. To integrate it into the block-sparse framework of InfLLM-V2, we max-pool InfiniGen's approximate attention scores along the key dimension with a kernel size equal to the block size, converting them to block-level scores. The resulting scores use the same model-specific budget allocation within 96 attended blocks and the same block-sparse attention kernel as Sparse and \method. Since the original InfiniGen does not accelerate prefill, we use InfLLM-V2 sparse prefill for InfiniGen and additionally build its partial key cache during prefill by projecting full post-RoPE keys onto a subset of dimensions for decode-time scoring.

\paragraph{HELMET.}
We evaluate at each model's native maximum (64K for MiniCPM4.1-8B, 32K for NOSA-8B) across all seven HELMET categories (recall, RAG, ICL, rerank, citation, long QA, summarization). Each category uses 100 test samples. Decoding is greedy ($\texttt{temperature}{=}0$, $\texttt{do\_sample}{=}\texttt{False}$). Generation length varies per category (50--200 tokens). For tasks originally designed to be judged by GPT-4, we use GPT-5.2 as the judge.

\paragraph{LongBench.}
We evaluate all 14 English tasks at the native maximum. Decoding is greedy with task-specific maximum generation lengths (32--512 tokens depending on the task).

\paragraph{RULER.}
We evaluate all 13 synthetic tasks with 50 examples per task. Default sequence length is the native maximum; extended evaluations use 32K, 64K, 96K, and 128K. Decoding is greedy with task-specific generation limits (30--128 tokens).

\paragraph{Reasoning (MATH-500, AIME 2024, AIME 2025).}
MATH-500 contains 500 problems; AIME 2024 and AIME 2025 each contain 30 problems. Unlike the other benchmarks, reasoning uses sampling ($\texttt{temperature}{=}0.9$, $\texttt{top\_p}{=}0.95$ for MiniCPM4.1-8B; $\texttt{temperature}{=}1.0$, $\texttt{top\_p}{=}1.0$ for NOSA-8B) with a long generation budget (up to 64K tokens for MiniCPM4.1-8B, 8K for NOSA-8B) to allow chain-of-thought reasoning. NOSA-8B does not support thinking mode, so we use a two-shot prompt that demonstrates step-by-step reasoning via in-context learning, following \citet{huang2025nosa}. Final answers are evaluated by a GPT-5.2 judge.

\paragraph{Multi-seed AIME evaluation.}
Each AIME dataset contains only 30 problems, making single-run scores sensitive to sampling variation. We therefore evaluate Dense, Sparse, InfiniGen, and \method\ on AIME 2024 and AIME 2025 with eight sampling seeds (1--8) for each model, using the settings above. Sparse and \method\ use paired seeds; the Dense and InfiniGen runs also share their per-problem sampling seeds. Table~\ref{tab:aime_multiseed} reports avg@8 $\pm$ standard deviation across per-seed accuracies, with one sampled answer per problem per run. Thus, avg@8 averages single-attempt accuracies rather than measuring whether any of eight answers is correct. Dense and InfiniGen standard deviations use the population convention over the eight runs.

We use these AIME means throughout for all four configurations, including the per-dataset scores, reasoning averages, and overall averages. Reasoning is the arithmetic mean of the MATH-500 score and the two AIME means, with rounding applied after aggregation. MATH-500 retains its single-run scores. The $(32,16)$ target-window ablation also retains single-run AIME scores; the $(2,1)$ variant is the main \method\ configuration and uses the eight-seed means.

Relative to Sparse, \method\ has a higher mean and lower seed-to-seed standard deviation in all four model--dataset comparisons. Dense has the highest AIME means on MiniCPM4.1-8B, while \method\ has the highest on NOSA-8B. Given the observed sampling variation, we interpret the Sparse--\method\ comparisons as supporting comparable accuracy and do not claim statistically significant gains from these eight runs.

\begin{table}[!htbp]
\centering
\small
\setlength{\tabcolsep}{8pt}
\caption{AIME accuracy (GPT-5.2 judge, \%), reported as avg@8 $\pm$ standard deviation over eight sampling seeds for every method. Each run evaluates all 30 problems per dataset.}
\label{tab:aime_multiseed}
\begin{tabular}{llrr}
\toprule
Model & Method & AIME 2024 & AIME 2025 \\
\midrule
\multirow{4}{*}{MiniCPM4.1-8B} & Dense & $84.6 \pm 4.1$ & $76.3 \pm 4.2$ \\
 & Sparse & $83.8 \pm 4.9$ & $72.1 \pm 7.9$ \\
 & InfiniGen & $82.9 \pm 3.5$ & $72.9 \pm 3.9$ \\
 & \method & $84.2 \pm 4.6$ & $75.4 \pm 5.6$ \\
\midrule
\multirow{4}{*}{NOSA-8B} & Dense & $55.8 \pm 7.0$ & $32.1 \pm 5.5$ \\
 & Sparse & $49.2 \pm 10.4$ & $30.0 \pm 9.4$ \\
 & InfiniGen & $54.6 \pm 9.1$ & $35.8 \pm 7.2$ \\
 & \method & $56.7 \pm 5.6$ & $36.2 \pm 8.4$ \\
\bottomrule
\end{tabular}
\end{table}

\paragraph{Efficiency.}
Throughput is measured on two GPU configurations: (1) NVIDIA H100 GPU with 80\,GB HBM3 memory and PCIe Gen5$\times$16 CPU-GPU connection, and (2) NVIDIA A100 GPU with 80\,GB HBM2e memory and PCIe Gen4$\times$16 CPU-GPU connection. Both systems have 2\,TB of CPU memory. For each throughput setting, we discard one warmup run and report the average of two measured runs. All methods in the 8B throughput evaluation are implemented on top of the high-throughput NOSI inference engine from \citet{huang2025nosa}. NOSI caches the top-$k$ KV blocks used by the previous decoding step for each layer by default, so each decode step only needs to fetch the newly selected KV blocks that were not in the previous step's selection. For offloading, pinned CPU memory stores the full KV history; the GPU retains compressed keys $\widetilde{\mathbf{K}}_l$ (${\approx}1/s_{C_1}$ the size of $\mathbf{K}_l$), initial/local blocks, and cached selected blocks. \method\ additionally keeps the full layer-0 KV on GPU because there is no prior Forecast to prefetch it. In the 32-layer GQA configuration, this costs about 1\,KB per token in FP16/BF16, or about 8\,GB at 128K context and batch 64 (1/32 of the full KV cache), explaining the earlier OOM at large context--batch settings. Layer 0 could instead be offloaded at the cost of a transfer without one-layer-ahead overlap.

\section{Additional Results}
\label{sec:appendix_additional}
\label{sec:appendix_detailed_accuracy}

\paragraph{Compression window ablation.}
The finer $(l_{C_1}{=}2, s_{C_1}{=}1)$ target compression window used during indexer training (Section~\ref{sec:indexer_training}) yields higher measured aggregate scores on both models (Table~\ref{tab:ablation_accuracy}). On MiniCPM4.1-8B, its RULER score is higher by +3.0, while HELMET and LongBench are nearly unchanged. On NOSA-8B, its HELMET, LongBench, and RULER scores are higher, including +2.5 on RULER, while the reasoning score is 54.8 versus 55.4 for the coarser window. The fine-window reasoning scores use eight-seed AIME means, whereas the coarse-window scores use single runs, so the reasoning differences should be interpreted cautiously. These long-context benchmark results motivate our use of the finer target window.

\begin{table*}[!htbp]
\centering
\small
\setlength{\tabcolsep}{4pt}
\caption{Ablation on the target compression window $(l_{C_1}, s_{C_1})$ for indexer training. The prediction window is fixed at $(32,16)$. Reasoning uses eight-seed AIME means for the $(2,1)$ variant and single-run scores for $(32,16)$; MATH-500 is single-run for both. \textit{Avg} is the arithmetic mean of the four benchmark families, computed before rounding. Best values per column are \textbf{bolded}.}
\label{tab:ablation_accuracy}
\resizebox{\textwidth}{!}{%
\begin{tabular}{l|rrrrr|rrrrr}
\toprule
 & \multicolumn{5}{c|}{MiniCPM4.1-8B} & \multicolumn{5}{c}{NOSA-8B} \\
Target window & HELMET & LongBench & RULER & Reasoning & Avg & HELMET & LongBench & RULER & Reasoning & Avg \\
\midrule
$(32,16)$ & 38.2 & \textbf{45.1} & 75.7 & 82.5 & 60.4 & 31.6 & 42.0 & 71.4 & \textbf{55.4} & 50.1 \\
$(2,1)$ & \textbf{38.3} & \textbf{45.1} & \textbf{78.7} & \textbf{85.7} & \textbf{61.9} & \textbf{33.4} & \textbf{42.3} & \textbf{73.9} & 54.8 & \textbf{51.1} \\
\bottomrule
\end{tabular}
}
\end{table*}

\paragraph{Prefetch CTA allocation ablation.}
Table~\ref{tab:prefetch_cta_sweep} sweeps the number of persistent-kernel CTAs on NVIDIA H100 GPU with MiniCPM4.1-8B at 32K. The optimal count generally grows with batch size. Using 16 CTAs is best at small batches (B8--B16), while 32 (or 64) CTAs are preferred at larger batches. Our adaptive heuristic uses 16 CTAs when batch size $< 32$ and 32 CTAs otherwise, matching or staying within 4\% of the best fixed configuration at every batch size. For the NVIDIA A100 GPU, a similar ablation yields an adaptive scheme of 16 CTAs when batch size $< 64$ and 32 CTAs otherwise.

\begin{table}[!htbp]
\centering
\small
\caption{Prefetch CTA-count sweep on NVIDIA H100 GPU with MiniCPM4.1-8B at 32K context. Best throughput at each batch size, including ties, is \textbf{bolded}.}
\label{tab:prefetch_cta_sweep}
\begin{tabular}{lrrrrrr}
\toprule
Method & B4 & B8 & B16 & B32 & B64 & B128 \\
\midrule
SparDA (8 CTAs) & 236.4 & 439.1 & 763.3 & 858.1 & 904.0 & 929.5 \\
SparDA (16 CTAs) & 235.7 & \textbf{471.9} & \textbf{918.5} & 1186.3 & 1255.9 & 1319.1 \\
SparDA (32 CTAs) & 233.5 & 471.5 & 813.1 & \textbf{1205.2} & 1328.0 & \textbf{1358.2} \\
SparDA (64 CTAs) & \textbf{237.0} & \textbf{471.9} & 737.3 & 1123.2 & \textbf{1380.2} & 1338.4 \\
\midrule
SparDA (adaptive) & 235.7 & \textbf{471.9} & \textbf{918.5} & \textbf{1205.2} & 1328.0 & \textbf{1358.2} \\
\bottomrule
\end{tabular}
\end{table}

\paragraph{A100 throughput results.}
Tables~\ref{tab:a100_prefill} and \ref{tab:a100_decode} report prefill and decode throughput on an NVIDIA A100 GPU, providing a comparison with the H100 results in the main text.

\begin{table*}[!htbp]
\centering
\small
\caption{Prefill throughput (tok/s) on NVIDIA A100 GPU at batch size 4. Sequence length swept from 32K to 128K. $\dagger$: no offload. Best value per column is \textbf{bolded}.}
\label{tab:a100_prefill}
\begin{tabular}{l|rrrr|rrrr}
\toprule
 & \multicolumn{4}{c|}{MiniCPM4.1-8B} & \multicolumn{4}{c}{NOSA-8B} \\
Method & 32K & 64K & 96K & 128K & 32K & 64K & 96K & 128K \\
\midrule
Dense$\dagger$ & \textbf{9531.5} & 6836.1 & 5260.2 & 4282.9 & \textbf{9576.2} & \textbf{6870.5} & 5313.5 & 4316.0 \\
Sparse$\dagger$ & 8002.2 & 7310.1 & 6776.8 & 6310.3 & 5833.2 & 5364.6 & 5022.4 & 4713.5 \\
Sparse & 8052.3 & 7311.0 & 6765.9 & 6299.2 & 5839.7 & 5365.0 & 5021.9 & 4717.1 \\
InfiniGen & 7985.5 & 7268.6 & 6740.0 & 6296.3 & 5815.4 & 5339.0 & 5005.2 & 4711.5 \\
SparDA & 8555.9 & \textbf{8191.3} & \textbf{7959.4} & \textbf{7756.8} & 6097.4 & 5817.9 & \textbf{5651.5} & \textbf{5477.5} \\
\bottomrule
\end{tabular}
\end{table*}

\begin{table*}[!htbp]
\centering
\small
\setlength{\tabcolsep}{3pt}
\caption{Decode throughput (tok/s) on NVIDIA A100 GPU. Batch size (B) swept from 4 to 128. $\dagger$: no offload. ``--'': OOM. Best value per column is \textbf{bolded}.}
\label{tab:a100_decode}
\resizebox{\textwidth}{!}{%
\begin{tabular}{cl|rrrrrr|rrrrrr}
\toprule
& & \multicolumn{6}{c|}{MiniCPM4.1-8B} & \multicolumn{6}{c}{NOSA-8B} \\
Context & Method & B4 & B8 & B16 & B32 & B64 & B128 & B4 & B8 & B16 & B32 & B64 & B128 \\
\midrule
 & Dense$\dagger$ & 160.9 & 219.7 & 268.8 & -- & -- & -- & \textbf{161.0} & 218.6 & 266.8 & -- & -- & -- \\
 & Sparse$\dagger$ & \textbf{171.4} & \textbf{348.6} & \textbf{688.7} & -- & -- & -- & 155.7 & 307.8 & \textbf{630.3} & -- & -- & -- \\
32K & Sparse & 165.6 & 276.1 & 405.9 & 515.7 & 597.9 & 649.2 & 158.5 & \textbf{317.8} & 572.7 & 828.2 & 1097.9 & 1113.3 \\
 & InfiniGen & 55.2 & 71.8 & 75.2 & 71.3 & 72.3 & 27.3 & 46.8 & 108.1 & 154.4 & 182.3 & 157.8 & 119.8 \\
 & SparDA & 160.4 & 316.4 & 568.0 & \textbf{649.2} & \textbf{668.9} & \textbf{691.9} & 148.1 & 292.4 & 591.4 & \textbf{988.6} & \textbf{1211.1} & \textbf{1218.1} \\
\midrule
 & Dense$\dagger$ & 109.6 & 134.8 & -- & -- & -- & -- & 109.5 & 133.7 & -- & -- & -- & -- \\
 & Sparse$\dagger$ & \textbf{173.2} & \textbf{344.6} & -- & -- & -- & -- & 157.4 & \textbf{310.3} & -- & -- & -- & -- \\
64K & Sparse & 147.3 & 236.7 & 345.6 & 443.9 & 516.1 & 556.6 & \textbf{158.5} & 284.6 & 479.8 & 704.8 & 911.8 & 861.3 \\
 & InfiniGen & 54.0 & 67.5 & 66.9 & 63.3 & 27.2 & -- & 63.3 & 100.1 & 133.5 & 138.1 & 62.5 & -- \\
 & SparDA & 160.3 & 316.3 & \textbf{457.8} & \textbf{573.7} & \textbf{593.1} & \textbf{578.4} & 146.9 & 292.8 & \textbf{577.3} & \textbf{887.8} & \textbf{1042.6} & \textbf{961.3} \\
\midrule
 & Dense$\dagger$ & 83.3 & 97.4 & -- & -- & -- & -- & 82.9 & 96.7 & -- & -- & -- & -- \\
 & Sparse$\dagger$ & 157.8 & 301.6 & -- & -- & -- & -- & \textbf{153.0} & 294.8 & -- & -- & -- & -- \\
96K & Sparse & 127.6 & 208.7 & 310.7 & 402.5 & 467.0 & \textbf{503.1} & 140.7 & 253.5 & 424.0 & 629.9 & 814.2 & \textbf{849.0} \\
 & InfiniGen & 50.8 & 62.0 & 57.5 & 53.2 & -- & -- & 61.8 & 97.0 & 124.7 & 122.6 & -- & -- \\
& SparDA & \textbf{158.7} & \textbf{306.6} & \textbf{436.0} & \textbf{522.6} & \textbf{529.4} & -- & 146.1 & \textbf{295.2} & \textbf{568.1} & \textbf{778.4} & \textbf{930.5} & -- \\
\midrule
 & Dense$\dagger$ & 67.2 & -- & -- & -- & -- & -- & 66.6 & 75.3 & -- & -- & -- & -- \\
 & Sparse$\dagger$ & 141.0 & -- & -- & -- & -- & -- & 136.5 & -- & -- & -- & -- & -- \\
128K & Sparse & 117.3 & 195.1 & 287.3 & 383.1 & 447.9 & -- & 127.4 & 232.4 & 395.6 & 585.4 & 795.9 & -- \\
 & InfiniGen & 49.6 & 62.3 & 58.1 & 41.3 & -- & -- & 62.8 & 95.7 & 70.3 & 62.2 & -- & -- \\
 & SparDA & \textbf{160.2} & \textbf{303.3} & \textbf{424.4} & \textbf{485.5} & \textbf{501.4} & -- & \textbf{148.4} & \textbf{298.2} & \textbf{526.3} & \textbf{691.3} & \textbf{876.9} & -- \\
\bottomrule
\end{tabular}
}
\end{table*}

On prefill, \method\ achieves up to 1.23$\times$ speedup over Sparse and 1.81$\times$ over Dense on MiniCPM4.1-8B at 128K, and up to 1.16$\times$ over Sparse on NOSA-8B. These gains are consistent with the H100 results, confirming that the Forecast indexer's reduced selection cost transfers across GPU generations.

On decode, SparDA achieves up to 1.55$\times$ speedup over Sparse on MiniCPM4.1-8B (128K, B8) and up to 1.34$\times$ on NOSA-8B (96K, B16), slightly lower than on H100. The same trends hold: speedups peak at middle batch sizes and NOSA-8B shows lower gains due to its reduced KV fetch traffic. InfiniGen's CPU-side gather bottleneck is even more pronounced on A100, with throughput degrading sharply at larger batch sizes.

\paragraph{Decode speedup breakdown.}
Table~\ref{tab:speedup_breakdown} isolates the two sources of \method's decode speedup on NVIDIA H100 GPU at 128K sequence length: reduced block-selection cost and overlapped CPU-to-GPU prefetch. \method\ (no prefetch) uses the Forecast indexer but fetches missing KV blocks synchronously as in Sparse, without asynchronous overlap. Even without prefetch, this variant already outperforms Sparse at B4 on both models because the Forecast indexer cuts block-selection latency. At this small batch, the prefetch pipeline adds slight overhead that makes the full \method\ marginally slower than the no-prefetch variant. As batch size grows, layer execution scales sub-linearly while KV transfer volume grows linearly. Prefetch overlap therefore becomes increasingly beneficial. By B16 the full \method\ pulls ahead of the no-prefetch variant on both models, and by B64 it is roughly 40\% faster. At B64, \method\ (no prefetch) drops below Sparse on both models because the block-selection savings diminish at large batches while the indexer projection cost remains, confirming that prefetch overlap is essential for \method's advantage at high batch sizes.

\begin{table}[!htbp]
\centering
\small
\setlength{\tabcolsep}{4pt}
\caption{Decode throughput (tok/s) breakdown on NVIDIA H100 GPU at 128K sequence length. Batch size (B) swept from 4 to 64. Best value per column is \textbf{bolded}.}
\label{tab:speedup_breakdown}
\begin{tabular}{l|rrrrr|rrrrr}
\toprule
 & \multicolumn{5}{c|}{MiniCPM4.1-8B} & \multicolumn{5}{c}{NOSA-8B} \\
Method & B4 & B8 & B16 & B32 & B64 & B4 & B8 & B16 & B32 & B64 \\
\midrule
Sparse & 167.8 & 279.5 & 447.9 & 618.6 & 788.9 & 173.2 & 285.4 & 529.2 & 898.7 & 1298.3 \\
SparDA (no prefetch) & \textbf{247.3} & 379.1 & 505.9 & 608.2 & 696.9 & \textbf{226.5} & \textbf{434.9} & 653.0 & 884.2 & 1063.8 \\
SparDA & 240.2 & \textbf{471.2} & \textbf{705.3} & \textbf{899.2} & \textbf{1000.1} & 219.0 & 399.3 & \textbf{735.0} & \textbf{1127.0} & \textbf{1463.3} \\
\bottomrule
\end{tabular}
\end{table}

\paragraph{Per-task accuracy breakdowns.}
Tables~\ref{tab:helmet_detail}--\ref{tab:reasoning_detail} report per-task scores for each benchmark family, complementing the aggregated results in Table~\ref{tab:main_accuracy}.

\begin{table*}[h]
\centering
\small
\caption{Per-category HELMET scores. Best among sparse methods (excluding Dense) is \textbf{bolded}.}
\label{tab:helmet_detail}
\begin{tabular}{l|rrrr|rrrr}
\toprule
& \multicolumn{4}{c|}{MiniCPM4.1-8B (64K)} & \multicolumn{4}{c}{NOSA-8B (32K)} \\
Category & Dense & Sparse & InfiniGen & SparDA & Dense & Sparse & InfiniGen & SparDA \\
\midrule
Recall & 84.7 & \textbf{67.8} & 37.8 & 61.5 & 77.9 & 40.9 & 19.6 & \textbf{45.6} \\
RAG & 55.1 & \textbf{54.7} & 51.4 & 54.6 & 51.4 & 49.1 & 47.4 & \textbf{49.4} \\
ICL & 84.0 & 82.2 & \textbf{82.3} & 81.9 & 64.3 & 68.4 & \textbf{69.1} & 67.9 \\
Citation & 2.5 & \textbf{3.5} & 3.0 & 3.2 & 3.4 & 3.2 & 2.9 & \textbf{3.9} \\
Rerank & 12.5 & 14.7 & 13.5 & \textbf{17.0} & 32.7 & 22.3 & 15.9 & \textbf{24.7} \\
LongQA & 31.8 & 30.9 & 30.2 & \textbf{32.1} & 28.3 & 26.9 & \textbf{27.4} & 26.0 \\
Summ & 20.9 & \textbf{18.2} & 16.4 & 18.0 & 17.0 & 14.8 & 14.6 & \textbf{16.5} \\
\midrule
Average & 41.7 & \textbf{38.9} & 33.5 & 38.3 & 39.3 & 32.2 & 28.1 & \textbf{33.4} \\
\bottomrule
\end{tabular}
\end{table*}

\begin{table}[H]
\centering
\small
\caption{Per-task LongBench scores. Best among sparse methods (excluding Dense) is \textbf{bolded}.}
\label{tab:longbench_detail}
\begin{tabular}{l|rrrr|rrrr}
\toprule
& \multicolumn{4}{c|}{MiniCPM4.1-8B} & \multicolumn{4}{c}{NOSA-8B} \\
Task & Dense & Sparse & InfiniGen & SparDA & Dense & Sparse & InfiniGen & SparDA \\
\midrule
2wikimqa & 33.1 & 35.2 & \textbf{35.4} & 34.2 & 37.8 & \textbf{36.8} & 35.5 & 35.5 \\
gov\_report & 28.5 & \textbf{28.2} & 27.8 & 28.0 & 32.8 & 33.3 & 32.9 & \textbf{33.6} \\
hotpotqa & 48.6 & 48.9 & \textbf{49.6} & 49.2 & 50.5 & 48.0 & 47.5 & \textbf{48.2} \\
multifieldqa\_en & 53.3 & 53.8 & 53.5 & \textbf{54.3} & 50.5 & 51.4 & 51.4 & \textbf{51.5} \\
musique & 23.7 & \textbf{25.4} & 24.6 & 25.2 & 25.0 & 23.4 & 21.0 & \textbf{24.9} \\
narrativeqa & 20.7 & 20.2 & 22.5 & \textbf{22.6} & 23.1 & \textbf{23.0} & 22.0 & 21.8 \\
passage\_count & 4.0 & \textbf{3.5} & \textbf{3.5} & \textbf{3.5} & 3.5 & 3.0 & \textbf{4.0} & 3.0 \\
passage\_retr. & 100.0 & \textbf{100.0} & \textbf{100.0} & \textbf{100.0} & 89.0 & \textbf{92.0} & 87.0 & 90.5 \\
qasper & 42.0 & \textbf{42.0} & 41.8 & 41.9 & 34.8 & \textbf{34.6} & 33.9 & 34.3 \\
qmsum & 24.2 & 23.6 & \textbf{24.0} & 23.7 & 24.8 & \textbf{25.0} & 24.5 & 24.7 \\
repobench-p & 50.6 & 50.9 & \textbf{51.2} & 50.0 & 53.9 & \textbf{54.1} & 53.8 & 53.6 \\
samsum & 39.8 & \textbf{40.8} & 39.8 & 40.6 & 6.9 & \textbf{6.1} & 6.0 & \textbf{6.1} \\
trec & 76.0 & \textbf{74.5} & \textbf{74.5} & \textbf{74.5} & 74.5 & 74.5 & 75.0 & \textbf{75.5} \\
triviaqa & 83.0 & \textbf{83.6} & \textbf{83.6} & \textbf{83.6} & 88.2 & \textbf{88.2} & 87.7 & \textbf{88.2} \\
\midrule
Average & 44.8 & 45.0 & \textbf{45.1} & \textbf{45.1} & 42.5 & \textbf{42.4} & 41.6 & 42.3 \\
\bottomrule
\end{tabular}
\end{table}

\begin{table}[H]
\centering
\small
\caption{Per-task RULER scores at the native maximum (64K for MiniCPM4.1-8B, 32K for NOSA-8B). Best among sparse methods (excluding Dense) is \textbf{bolded}.}
\label{tab:ruler_detail}
\begin{tabular}{l|rrrr|rrrr}
\toprule
& \multicolumn{4}{c|}{MiniCPM4.1-8B (64K)} & \multicolumn{4}{c}{NOSA-8B (32K)} \\
Task & Dense & Sparse & InfiniGen & SparDA & Dense & Sparse & InfiniGen & SparDA \\
\midrule
niah\_single\_1 & 100.0 & \textbf{100.0} & \textbf{100.0} & \textbf{100.0} & 100.0 & \textbf{100.0} & \textbf{100.0} & \textbf{100.0} \\
niah\_single\_2 & 100.0 & \textbf{100.0} & \textbf{100.0} & \textbf{100.0} & 100.0 & \textbf{100.0} & \textbf{100.0} & \textbf{100.0} \\
niah\_single\_3 & 100.0 & \textbf{100.0} & 86.0 & \textbf{100.0} & 100.0 & \textbf{100.0} & 98.0 & \textbf{100.0} \\
niah\_multikey\_1 & 98.0 & 94.0 & 96.0 & \textbf{100.0} & 100.0 & 86.0 & 70.0 & \textbf{90.0} \\
niah\_multikey\_2 & 98.0 & \textbf{80.0} & 44.0 & 78.0 & 98.0 & 50.0 & 18.0 & \textbf{52.0} \\
niah\_multikey\_3 & 98.0 & \textbf{62.0} & 28.0 & 56.0 & 90.0 & 20.0 & 8.0 & \textbf{28.0} \\
niah\_multivalue & 94.5 & 95.5 & 90.5 & \textbf{96.5} & 94.0 & 92.0 & 85.0 & \textbf{95.0} \\
niah\_multiquery & 99.0 & 97.5 & 87.5 & \textbf{98.5} & 95.0 & 90.0 & 71.5 & \textbf{95.0} \\
vt & 54.8 & 74.4 & 73.2 & \textbf{74.8} & 98.8 & 97.2 & 97.2 & \textbf{98.4} \\
cwe & 44.4 & 11.0 & \textbf{23.2} & 12.4 & 47.6 & 31.6 & \textbf{37.8} & 34.6 \\
fwe & 92.0 & \textbf{94.0} & 79.3 & 92.7 & 95.3 & 89.3 & \textbf{90.7} & 86.0 \\
qa\_1 & 80.0 & 64.0 & 44.0 & \textbf{70.0} & 56.0 & 38.0 & 32.0 & \textbf{40.0} \\
qa\_2 & 50.0 & \textbf{44.0} & 38.0 & \textbf{44.0} & 46.0 & \textbf{44.0} & 40.0 & 42.0 \\
\midrule
Average & 85.3 & 78.2 & 68.4 & \textbf{78.7} & 86.2 & 72.2 & 65.2 & \textbf{73.9} \\
\bottomrule
\end{tabular}
\end{table}

\begin{table}[H]
\centering
\small
\setlength{\tabcolsep}{5pt}
\caption{Per-dataset reasoning accuracy (GPT-5.2 judge, \%). All AIME entries are eight-seed means $\pm$ standard deviations, as in Table~\ref{tab:aime_multiseed}; MATH-500 is single-run. \textit{Average} is the mean of the three datasets, computed before rounding. Best among sparse methods (excluding Dense) is \textbf{bolded}.}
\label{tab:reasoning_detail}
\begin{tabular}{ll|rrrr}
\toprule
Model & Dataset & Dense & Sparse & InfiniGen & SparDA \\
\midrule
\multirow{4}{*}{MiniCPM4.1-8B} & MATH-500 & 96.8 & 97.6 & \textbf{97.8} & 97.4 \\
 & AIME 2024 & $84.6 \pm 4.1$ & $83.8 \pm 4.9$ & $82.9 \pm 3.5$ & $\mathbf{84.2} \pm 4.6$ \\
 & AIME 2025 & $76.3 \pm 4.2$ & $72.1 \pm 7.9$ & $72.9 \pm 3.9$ & $\mathbf{75.4} \pm 5.6$ \\
 & Average & 85.9 & 84.5 & 84.5 & \textbf{85.7} \\
\midrule
\multirow{4}{*}{NOSA-8B} & MATH-500 & 68.2 & 72.2 & \textbf{72.8} & 71.6 \\
 & AIME 2024 & $55.8 \pm 7.0$ & $49.2 \pm 10.4$ & $54.6 \pm 9.1$ & $\mathbf{56.7} \pm 5.6$ \\
 & AIME 2025 & $32.1 \pm 5.5$ & $30.0 \pm 9.4$ & $35.8 \pm 7.2$ & $\mathbf{36.2} \pm 8.4$ \\
 & Average & 52.0 & 50.5 & 54.4 & \textbf{54.8} \\
\bottomrule
\end{tabular}
\end{table}

\subsection{Retrieval Failure Analysis}
\label{sec:appendix_retrieval}

\paragraph{Diagnostic evaluation.}
We instrument decode-time block selection on MiniCPM4.1-8B at 64K context across all 13 RULER tasks, using 20 examples per task. This is a separate diagnostic evaluation from the 50-example-per-task results in Table~\ref{tab:ruler_detail}. Sparse selects with the actual query and coarse compressed keys, while \method\ uses the Forecast. The reference is the finer attention-derived training target described in Section~\ref{sec:indexer_training}; for this analysis, its scores are pooled to 64-token blocks before selecting the reference top-$k$ set.

\paragraph{Selection agreement and answer coverage.}
\emph{Target recall} is the fraction of target-selected blocks retained by a selector. \emph{Needle recall} is the fraction of answer-containing blocks retained. Table~\ref{tab:retrieval_failure} reports target recall for Sparse and \method, and needle recall for the target and both selectors. These metrics distinguish agreement with the training reference from coverage of the answer locations.

\begin{table}[!htbp]
\centering
\small
\setlength{\tabcolsep}{4pt}
\caption{Retrieval diagnostics on MiniCPM4.1-8B at 64K, with 20 examples per RULER task. Task scores are percentages; recalls are proportions. The target is the fine attention-derived training reference, evaluated at the same block granularity as the two selectors. These diagnostic scores are reported separately from Table~\ref{tab:ruler_detail}.}
\label{tab:retrieval_failure}
\begin{tabular}{l|rrr|rr|rrr}
\toprule
 & \multicolumn{3}{c|}{Task score (\%)} & \multicolumn{2}{c|}{Target recall} & \multicolumn{3}{c}{Needle recall} \\
Task & Dense & Sparse & SparDA & Sparse & SparDA & Target & Sparse & SparDA \\
\midrule
niah\_single\_1 & 100 & 100 & 100 & 0.68 & 0.49 & 0.89 & 0.84 & 0.86 \\
niah\_single\_2 & 100 & 100 & 100 & 0.57 & 0.49 & 0.76 & 0.66 & 0.70 \\
niah\_single\_3 & 100 & 100 & 100 & 0.61 & 0.51 & 0.68 & 0.64 & 0.64 \\
niah\_multikey\_1 & 100 & 95 & 100 & 0.57 & 0.49 & 0.71 & 0.60 & 0.64 \\
niah\_multikey\_2 & 100 & 85 & 85 & 0.57 & 0.48 & 0.30 & 0.26 & 0.23 \\
niah\_multikey\_3 & 100 & 75 & 65 & 0.56 & 0.47 & 0.28 & 0.27 & 0.25 \\
niah\_multiquery & 99 & 100 & 99 & 0.59 & 0.50 & 0.56 & 0.43 & 0.50 \\
niah\_multivalue & 90 & 94 & 95 & 0.58 & 0.50 & 0.62 & 0.49 & 0.56 \\
vt & 52 & 69 & 66 & 0.68 & 0.54 & 0.83 & 0.68 & 0.73 \\
qa\_1 & 80 & 65 & 70 & 0.58 & 0.50 & 0.26 & 0.22 & 0.23 \\
qa\_2 & 40 & 40 & 40 & 0.56 & 0.48 & 0.29 & 0.26 & 0.29 \\
cwe & 41 & 11 & 11 & 0.55 & 0.47 & 0.08 & 0.07 & 0.07 \\
fwe & 92 & 93 & 93 & 0.52 & 0.44 & 0.18 & 0.14 & 0.20 \\
\bottomrule
\end{tabular}
\end{table}

\paragraph{Shared failures on harder retrieval tasks.}
Across multikey\_1--3, \method's target recall changes little (0.49 to 0.47), while the target's own needle recall falls from 0.71 to 0.28. Sparse and \method\ show similar declines in needle recall, from 0.60 to 0.27 and from 0.64 to 0.25, respectively. Their task scores also fall, while Dense stays at 100. On cwe, even the target retains only 0.08 of the answer-containing blocks; Sparse and \method\ both retain 0.07 and score 11, versus 41 for Dense. These observations suggest a shared difficulty in retaining relevant blocks that is not captured by Forecast-to-target agreement alone.

\paragraph{Forecast-specific differences and interpretation.}
\method\ has lower target recall than Sparse on all 13 tasks, yet equal or higher needle recall on 11. Its two needle-recall deficits occur on multikey\_2 and multikey\_3 (0.03 and 0.02). Accuracy is tied on multikey\_2, but \method\ scores 65 versus 75 on multikey\_3, a difference of two answers in this 20-example evaluation. This leaves room for an additional Forecast-selection cost on the hardest multikey task. Needle recall is also not a complete predictor of task accuracy. On vt, \method\ retains more answer-containing blocks (0.73 versus 0.68) but scores lower (66 versus 69). The study identifies retrieval failure patterns; it does not isolate the causal contribution of Forecast error.

\FloatBarrier

\section{Architectural Transfer to Token-Level DSA}
\label{sec:appendix_glm_transfer}

\paragraph{GLM-5.1 adaptation.}
GLM-5.1 is a 754B mixture-of-experts model with native token-level DSA~\citep{zai2026glm51,glm5team2026glm5}. We freeze the backbone and train only the Forecast projections at 64K sequence length, using KL distillation against the native DSA indexer's selection distribution. Thus, the teacher is the DSA indexer, rather than the full attention query used by the block-sparse backbones. The Forecast retains the native indexer's 32 scoring heads with head dimension 128. This indexer is already smaller than the model's 64-head attention module. Reducing the Forecast head count further degraded accuracy. We therefore do not obtain the prefill selection-compute savings of the compact GQA indexer. In the GLM-5.1 adaptation, Forecast predicts the next layer's token selection. The baseline uses the native DSA selector.

\paragraph{Evaluation settings.}
Both configurations run in SGLang~\citep{zheng2024sglang} on four nodes with eight NVIDIA H100 GPUs each (32 GPUs total). Tensor parallelism is 8, pipeline parallelism is 4, and expert parallelism is 8 within the same per-node GPU groups. RULER, HELMET, and LongBench use the 64K setting. RULER evaluates 50 examples for each of its 13 tasks; LongBench includes the same 14 tasks as the 8B evaluation, and HELMET reports seven category averages. GPT-5.2 is used for HELMET metrics requiring an LLM judge and for the reasoning suite. Reasoning uses a maximum generation budget of 32,768 tokens, temperature 0.9, and top-$p$ 0.95. MATH-500 is evaluated once; AIME 2024 and 2025 each use eight sampling seeds (1--8), shared between native DSA and \method, with one sampled answer per problem per seed. We report the mean and standard deviation of per-seed AIME accuracies, as in Appendix~\ref{sec:appendix_eval}. These GLM experiments evaluate accuracy.

\paragraph{Results.}
Tables~\ref{tab:glm_helmet_detail}--\ref{tab:glm_reasoning_detail} provide the detailed scores, complementing the benchmark-family averages in Table~\ref{tab:glm_accuracy} in Section~\ref{sec:glm_transfer}. The positive AIME mean differences are within the reported seed-to-seed variation and are not presented as statistically significant gains. Family and overall averages are computed before rounding; detailed scores are displayed to one decimal place.

\begin{table}[!htbp]
\centering
\small
\caption{GLM-5.1 HELMET scores by category at 64K.}
\label{tab:glm_helmet_detail}
\begin{tabular}{l|rr}
\toprule
Category & Native DSA & SparDA \\
\midrule
Recall & 100.0 & 99.8 \\
RAG & 75.6 & 75.6 \\
ICL & 90.2 & 90.8 \\
Citation & 52.4 & 51.3 \\
Rerank & 72.3 & 69.7 \\
LongQA & 62.0 & 62.5 \\
Summ & 45.3 & 44.4 \\
\midrule
Average & 71.1 & 70.6 \\
\bottomrule
\end{tabular}
\end{table}

\begin{table}[!htbp]
\centering
\small
\caption{GLM-5.1 LongBench scores by task at the 64K setting.}
\label{tab:glm_longbench_detail}
\begin{tabular}{l|rr}
\toprule
Task & Native DSA & SparDA \\
\midrule
2wikimqa & 75.3 & 74.0 \\
gov\_report & 31.6 & 31.6 \\
hotpotqa & 68.3 & 67.6 \\
multifieldqa\_en & 54.6 & 54.3 \\
musique & 56.4 & 56.5 \\
narrativeqa & 34.0 & 34.2 \\
passage\_count & 1.7 & 1.0 \\
passage\_retr. & 100.0 & 100.0 \\
qasper & 49.6 & 49.8 \\
qmsum & 23.4 & 23.2 \\
repobench-p & 67.3 & 68.4 \\
samsum & 39.2 & 39.6 \\
trec & 82.5 & 82.0 \\
triviaqa & 93.5 & 93.4 \\
\midrule
Average & 55.5 & 55.4 \\
\bottomrule
\end{tabular}
\end{table}

\begin{table}[!htbp]
\centering
\small
\caption{GLM-5.1 RULER scores by task at 64K.}
\label{tab:glm_ruler_detail}
\begin{tabular}{l|rr}
\toprule
Task & Native DSA & SparDA \\
\midrule
niah\_single\_1 & 100.0 & 100.0 \\
niah\_single\_2 & 100.0 & 100.0 \\
niah\_single\_3 & 100.0 & 100.0 \\
niah\_multikey\_1 & 100.0 & 100.0 \\
niah\_multikey\_2 & 100.0 & 100.0 \\
niah\_multikey\_3 & 100.0 & 100.0 \\
niah\_multivalue & 100.0 & 99.5 \\
niah\_multiquery & 100.0 & 100.0 \\
vt & 100.0 & 100.0 \\
cwe & 100.0 & 100.0 \\
fwe & 98.0 & 98.0 \\
qa\_1 & 84.0 & 84.0 \\
qa\_2 & 72.0 & 72.0 \\
\midrule
Average & 96.5 & 96.4 \\
\bottomrule
\end{tabular}
\end{table}

\begin{table}[!htbp]
\centering
\small
\caption{GLM-5.1 reasoning accuracy (GPT-5.2 judge, \%) with a 32K generation budget. AIME entries are eight-seed means $\pm$ standard deviations; MATH-500 is single-run. \textit{Average} is the mean of the three datasets.}
\label{tab:glm_reasoning_detail}
\begin{tabular}{l|rr}
\toprule
Dataset & Native DSA & SparDA \\
\midrule
MATH-500 & 99.6 & 99.8 \\
AIME 2024 & $98.8 \pm 1.7$ & $99.2 \pm 1.5$ \\
AIME 2025 & $93.8 \pm 3.8$ & $96.2 \pm 3.3$ \\
\midrule
Average & 97.4 & 98.4 \\
\bottomrule
\end{tabular}
\end{table}

\FloatBarrier

\section{Broader Impact}
\label{sec:appendix_broader_impact}

\method\ is an inference-efficiency technique that reduces the cost of serving long-context language models. By lowering the hardware required for a given throughput target, it could make long-context LLM capabilities more accessible to researchers and practitioners with limited GPU budgets. The same cost reduction can also lower the barrier to generating harmful content at scale. We do not introduce new model capabilities or training data, so the misuse surface is inherited from the underlying models rather than created by this work. We therefore encourage practitioners to apply existing safety filters, monitoring, and deployment controls when serving models accelerated by \method\ or similar techniques.

\FloatBarrier

\section{Licenses and Copyright for External Assets}
\label{sec:appendix_licenses}

Table~\ref{tab:licenses} lists the external assets used in this work and their reported licenses or copyright information. For AIME 2024 and AIME 2025, MAA denotes the Mathematical Association of America, which holds copyright in the original competition problems; the benchmark names link to the dataset sources.

\begin{table}[h]
\centering
\small
\caption{License and copyright information for external assets used in this paper.}
\label{tab:licenses}
\begin{tabular}{lll}
\toprule
Asset & Type & License / copyright \\
\midrule
MiniCPM4.1-8B & Model & Apache 2.0 \\
NOSA-8B & Model & Apache 2.0 \\
GLM-5.1 & Model & MIT \\
ProLong-64K & Dataset & MIT \\
HELMET & Benchmark & MIT \\
LongBench & Benchmark & MIT \\
RULER & Benchmark & Apache 2.0 \\
MATH-500 & Benchmark & MIT \\
\href{https://huggingface.co/datasets/math-ai/aime24}{AIME 2024} & Benchmark & MAA copyright \\
\href{https://huggingface.co/datasets/math-ai/aime25}{AIME 2025} & Benchmark & MAA copyright \\
NOSI & Library & MIT \\
SGLang & Library & Apache 2.0 \\
Triton & Library & MIT \\
PyTorch & Library & BSD-3-Clause \\
\bottomrule
\end{tabular}
\end{table}

\end{document}